\documentclass[11pt]{article}

\usepackage[preprint]{acl}

\usepackage{times}
\usepackage{latexsym}
\usepackage{stfloats}
\usepackage{amssymb}
\usepackage[T1]{fontenc}
\usepackage[utf8]{inputenc}

\usepackage{microtype}

\usepackage{inconsolata}

\usepackage{graphicx}

\usepackage{subfigure}
\usepackage{booktabs} % for professional tables
\usepackage{multirow}
\usepackage{makecell}
\usepackage{threeparttable}
\usepackage[dvipsnames]{xcolor}
\usepackage{xcolor}
\usepackage{xspace}
\usepackage{dsfont}
\usepackage{pifont}

\usepackage{algorithm}
\usepackage{algorithmic}

\usepackage{tcolorbox}
\usepackage{amsmath}
\usepackage{colortbl}
\usepackage{fancyvrb}
\tcbuselibrary{listings, breakable}
\usepackage{enumitem}

\definecolor{tablegray}{gray}{0.9}

\newtcolorbox[auto counter, number freestyle={\noexpand\arabic{\tcbcounter}}]{promptcolorbox}[3][]{%
    fonttitle=\bfseries,
    colframe=blue!50!black, 
    colback=blue!5, 
    title=#2~\thetcbcounter: #3,
    #1
}

\newtcolorbox[auto counter, number freestyle={\noexpand\arabic{\tcbcounter}}]{instructioncolorbox}[3][]{%
    fonttitle=\bfseries,
    title=#2~\thetcbcounter: #3,
    #1
}

\usepackage{hyperref}

\newcommand{\Ours}{\textsc{Online Adaptation to Continual Knowledge Streams}\xspace}
\newcommand{\ours}{\textsc{OAKS}\xspace}
\newcommand{\Babi}{\textsc{OAKS}-BABI\xspace}
\newcommand{\Novels}{\textsc{OAKS}-Novel\xspace}
\newcommand{\babi}{\textsc{OAKS}-B\xspace}
\newcommand{\novels}{\textsc{OAKS}-N\xspace}

\newcommand{\cmark}{\ding{51}}
\newcommand{\xmark}{\ding{55}}
\newcommand{\greencmark}{\textcolor{ForestGreen}{\ding{51}}}
\newcommand{\redxmark}{\textcolor{red}{\ding{55}}}

\title{
Can Large Language Models Keep Up? \\ Benchmarking Online Adaptation to Continual Knowledge Streams}

\author{
 \textbf{Jiyeon Kim\textsuperscript{*1}},
 \textbf{Hyunji Lee\textsuperscript{*2}},
 \textbf{Dylan Zhou\textsuperscript{*3}},
 \textbf{Sue Hyun Park\textsuperscript{4}},
\\
 \textbf{Seunghyun Yoon\textsuperscript{5}},
 \textbf{Trung Bui\textsuperscript{5}},
 \textbf{Franck Dernoncourt\textsuperscript{5}},
 \textbf{Sungmin Cha\textsuperscript{6}},
 \textbf{Minjoon Seo\textsuperscript{1}}
\\
\\
 \textsuperscript{1}KAIST AI,
 \textsuperscript{2}UNC Chapel Hill,
 \textsuperscript{3}Google,
\\
 \textsuperscript{4}KRAFTON,
 \textsuperscript{5}Adobe Research,
 \textsuperscript{6}New York University
\\
 \small{
    \textsuperscript{*}Equal Contribution
    }
\\
 \small{
   \textbf{Correspondence:} \href{mailto:jiyeon.kim@kaist.ac.kr}{jiyeon.kim@kaist.ac.kr}, \href{mailto:hyunjil@cs.unc.edu}{hyunjil@cs.unc.edu}, \href{mailto:dylanzhou@google.com}{dylanzhou@google.com}
 }
}

\begin{document}
\maketitle
\begin{abstract}
Large language models operating in dynamic real-world contexts often encounter knowledge that evolves continuously or emerges incrementally.
To remain accurate and effective, models must adapt to newly arriving information on the fly.
We introduce \Ours~(\ours) to evaluate this capability, establishing a benchmark for \textit{online adaptation over streaming, continually updating knowledge}.
Specifically, each model is evaluated at every time interval using the same set of questions, allowing us to assess whether it can track and reason over such fine-grained knowledge dynamics across time.
To support this setting, we present two datasets: \Babi and \Novels, where individual facts evolve multiple times across context chunks. 
These datasets include dense annotations to measure whether models track changes accurately.
Evaluating 14 models with varied inference approaches, we observe significant limitations in current methodologies. Both state-of-the-art models and agentic memory systems fail to adapt robustly on \ours, demonstrating delays in state-tracking and susceptibility to distraction within streaming environments.\footnote{
Project page: \href{https://github.com/kaistAI/OAKS}{https://github.com/kaistAI/OAKS}
}

\end{abstract}

\section{Introduction}

\begin{figure*}[t!]
    \centering
    \includegraphics[width=0.85\linewidth]{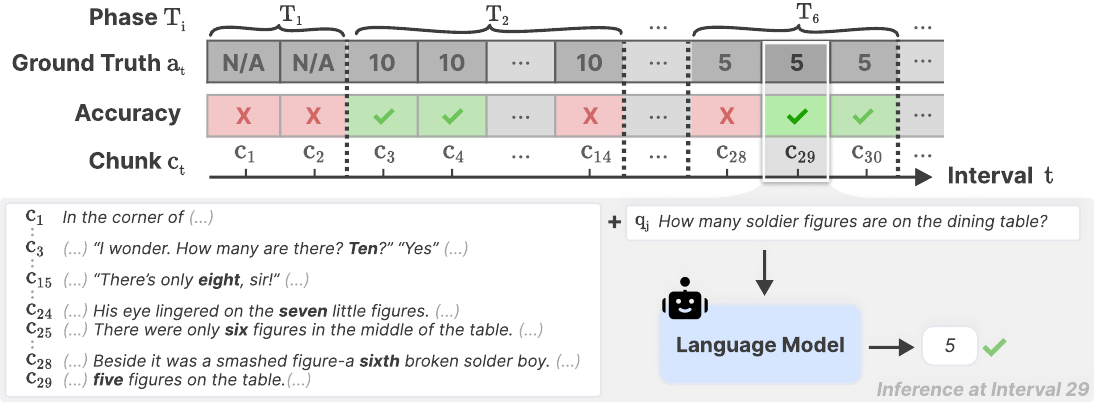}
    \caption{
    Overview of \Ours. 
    At each time interval $t$, a new context chunk $c_t$ is streamed, and the model is queried with context accumulated up to $t$ and a question $q_j$. 
    Performance is calculated as average accuracy across all intervals by comparing predictions with the ground truth answer $a_t$, then averaging over all questions. 
    A Phase $T_i$ denotes a contiguous set of chunks sharing the same ground truth answers. 
    Answer options are limited to \novels, as \babi uses an open-ended format.
    }
    \label{fig:overall}
\end{figure*}

In real-world settings, knowledge is inherently dynamic, evolving continuously and emerging incrementally.
Consequently, LLM-based systems operating as conversational assistants or embodied agents must adapt to information that changes sequentially over time on the fly~\citep{yu2025memagent, kim2024exploring, zheng2025lifelong}.
For example, assistants receive user context gradually during dialogue~\citep{maharana-etal-2024-evaluating, wu2025longmemeval}, and robots encounter new properties of their environments during exploration~\citep{majumder2023clin, kim2024online}. If the information updates are not integrated in real time, model predictions risk becoming outdated or even unsafe. However, current benchmarks primarily target static knowledge or offline tasks, failing to adequately evaluate online adaptation in dynamic settings.

To fill this void, we introduce \Ours~(\ours), a benchmark designed to evaluate models in an \textit{online adaptation} setting over \textit{streaming, continually updating knowledge}.\footnote{
\textit{Streaming} refers to a dataset characteristic where facts arrive sequentially over time. \textit{Online adaptation} refers to an inference setting in which a model adapts to the most up-to-date information on the fly.}
\ours synthesizes continual knowledge learning~\citep{liska2022streamingqa, jang2022towards} and online adaptation~\citep{lin2024streamingbench, hu2023meta}; facts arrive sequentially and may supersede or contradict prior information, necessitating that models dynamically revise their knowledge state.
Distinct from prior work, \ours evaluates whether models can track factual updates, maintain consistency, and generalize when reasoning over long-horizon streams characterized by frequent, fine-grained state changes.
To the best of our knowledge, \ours is the first benchmark to unify these two paradigms, supporting both large-scale fine-grained knowledge adaptation and stepwise online evaluation over streaming knowledge.

To evaluate models on \ours, we introduce new datasets: \Babi~(\babi), a synthetic dataset derived from the BABILong~benchmark \citep{kuratov2024babilong}, and \Novels~(\novels), a human-curated dataset sourced from literary texts.
As shown in Figure~\ref{fig:overall}, the datasets consist of multiple context chunks $c_t$, where facts evolve dynamically over time intervals $t$.
For each question, we annotate answers at each time interval based on all knowledge accumulated up to that point, capturing \textit{answer transitions} triggered by new information. 
This setup explicitly evaluates a model’s ability to track when and how answers change in an online setting.
During evaluation, the model is asked the same set of questions at each interval, with access to the context up to that point (e.g., chunks $c_1$ to $c_{29}$ when evaluating at interval 29, as shown in the figure).
Performance is measured by interval-level accuracy, reflecting whether the model maintains the correct state at each specific moment in time.

We conduct an extensive analysis of 14 state-of-the-art models on \babi and \novels, including strong closed-sourced model Gemini 3~\citep{gemini3} and various sizes of Qwen3~\citep{yang2025qwen3}, utilizing diverse strategies for constructing accumulated context.
Our results show that models struggle with \ours, achieving an average accuracy of 39.4\% on \babi and 57.5\% on \novels. 
We find that models particularly degrade with \textit{frequently updating} knowledge, with accuracies dropping to 33.3\% on \babi and 53.0\% on \novels. 
While inference-time scaling enhances performance on complex reasoning tasks, its gains are limited when tracking states that undergo frequent updates.
Across a range of context construction strategies, including agentic memory systems, naive retrieval augmented generation~(RAG), and recency-based approaches, we observe that \ours remains challenging across all settings. 
Among these methods, RAG tends to show robust performance, whereas agentic memory approaches outperform simple RAG when frequent knowledge updates exist.

Leveraging the online nature of \ours, which queries the same question at multiple points within a streaming context, we conduct a fine-grained analysis of models' knowledge tracking behavior. 
We find that activating ``thinking mode'' enhances both adaptability and stability, leading to higher overall accuracy. 
However, distinct failure modes emerge across different models: some tend to over-update, changing predictions unnecessarily, while others under-update, displaying inertia even when the underlying state shifts.
Our intra-phase analysis further reveals that even when models achieve comparable accuracy, the underlying causes of error diverge based on their state-transition behaviors: over-updating models tend to capture the correct phase but suffer from frequent distractions within it, whereas under-updating models are prone to missing entire phases. 
Finally, we observe that performance degrades on questions requiring reasoning over multiple context chunks and at later time intervals as context length increases.

\section{Related Work}
\paragraph{Continual Knowledge Learning and State Tracking Benchmarks}
Real-world knowledge is dynamic, motivating benchmarks that evaluate language models under settings where knowledge evolves continually~\citep{kim2024carpe, jang2022towards, liska2022streamingqa}. 
However, existing benchmarks typically involve a small number of knowledge updates or focus on divergent fact updates rather than the same underlying fact. 
Our work addresses this by tracking fine-grained continual updates. 
\ours is also related to state-tracking benchmarks, which study how models maintain and update evolving states~\citep{kim2023entity, niu2024enhancing}. 
While prior work focuses on short-term, structured states such as dialogue slots~\citep{budzianowski2020multiwozlargescalemultidomain, lee2019sumbt}, \ours instead focuses on open-ended knowledge states over long horizons, evaluating whether models can maintain temporal consistency over a continuous stream of updates in an online setting.\footnote{Due to length constraint, full list of related works is in Appendix~\ref{app:related_works}}

\paragraph{Online Adaptation to Streaming Inputs}
Prior research on online or lifelong learning in language has focused on self-evolving, lifelong agents that acquire new capabilities or task-level skills over time~\citep{zheng2025lifelongagentbench, wei2025evo}, or on tracking synthetic facts with limited update complexity~\citep{wu2023online, lyu2025facttrack}. 
In contrast, we focus on fine-grained knowledge updates, a setting where minor failures can easily compound into significant errors over time.
Inspired by real-time understanding frameworks in the video domain~\citep{lin2024streamingbench, niu2025ovo}, we introduce \ours to explicitly evaluate realistic streaming knowledge updates at the granularity of individual facts in an online setting.

\section{\Ours}

In this paper, we present \Ours~(\ours), a benchmark that simulates real-world scenarios in which language models encounter streaming knowledge and must adapt to incrementally revealed updates online. 
In Section~\ref{sec3: dataset}, we introduce new datasets, and in Section~\ref{sec3: metric}, we describe the evaluation setup and metric.

\begin{table}[t!]
    \fontsize{8}{10} \selectfont
    \begin{tabular}{l|crcc}
    \toprule
    Dataset & Unit & \#Steps & U/Q  & Fully Annot. \\
    \midrule
    EvolvingQA &Corpus & 6+ & 2.0 & \xmark \\
    StreamingQA & Corpus & 4 & 2.0 & \xmark \\
    StreamingBench & Frame & 5 & 1.0 & \xmark \\
    FactTrack& Fact & 39 & 2.0 & \xmark \\
    MultiWOZ& Turn & 14 & 1.0 & \cmark  \\
    BABILong& Full & 1 & 1.0 & \xmark \\
    \midrule
    \babi & Chunk & 65 & \textbf{4.7} & \cmark \\
    \novels& Chunk & 78 & \textbf{4.7} & \cmark \\
    \bottomrule
    \end{tabular}
    \centering
    \caption{Comparison between our dataset with prior datasets along four axes: \textit{Unit} = size of each data unit, \textit{\#Steps} = number of evaluation steps, \textit{U/Q} = average number of updates (\textit{U}) per question (\textit{Q}), \textit{Fully Annotated (Fully Annot.)} = whether all questions are annotated for each unit. 
    }
    \label{table: dataset_compare}
\end{table}

\subsection{Dataset} \label{sec3: dataset}
To evaluate model performance on \ours, we present two datasets: \Babi~(\babi), derived from the BABILong benchmark~\citep{kuratov2024babilong}, and \Novels~(\novels), sourced from full-length literary novels.
Both datasets feature sequential context chunks containing knowledge updates, with questions explicitly curated to target facts that evolve over time.
Formally, each data consists of a set of $Q$ questions $\{q_j\}_{j=1}^Q$ and an ordered sequence of $C$ context chunks $\{c_i\}_{i=1}^C$. At each time interval $t$, a new chunk $c_t$ (spanning 2k tokens) is revealed. For each question $q_j$, we provide a ground-truth answer $a_{j,t}$ and supporting evidence $e_{j,t}$, representing the valid knowledge state conditioned on the cumulative history $\{c_i\}_{i=1}^{t}$.

As detailed in Table~\ref{table: dataset_compare}, unlike prior continual knowledge benchmarks such as EvolvingQA~\citep{kim2024carpe} and StreamingQA~\citep{liska2022streamingqa}, our dataset is composed of granular context increments distributed over a large number of time steps. This structure, coupled with questions that track multiple state transitions, enables stepwise online evaluation.
Furthermore, compared to general online adaptation~(StreamingBench~\citep{lin2024streamingbench}) or fact-tracking benchmarks~(FactTrack~\citep{lyu2025facttrack}, MultiWOZ~\citep{budzianowski2020multiwozlargescalemultidomain}), we focus specifically on the evolution of semantic knowledge, facilitating a more fine-grained assessment of knowledge adaptation capabilities.
Since each question targets a different aspect of knowledge from the context, the timing and frequency of phase transitions vary across questions. 
To evaluate model robustness under varying degrees of change, we stratify the dataset into three subsets, \textit{Sparse, Moderate, Frequent}, based on the frequency of answer changes for each question. 

Additional details on dataset construction, filtering, human annotation, refinement, and statistics on \babi and \novels can be found in Appendix~\ref{app:dataset}.

\paragraph{\babi}
\babi repurposes context from BABILong~\citep{kuratov2024babilong}, shifting the focus from static fact retrieval to dynamic knowledge tracking and reasoning. 
To achieve this, we reformulated the dataset in two ways: 
(1) generated \textit{new questions} that focus on state changes, ensuring that answers vary depending on the part of the context being considered and require synthesizing or reasoning over multiple facts; 
(2) annotated \textit{answers at every time interval} explicitly tracking how facts evolve over the stream.
The resulting dataset includes four question types: tracking, counting, bridge, and comparison.
Tracking questions focus on understanding frequent fact updates; counting, bridge, and comparison questions require aggregation and reasoning over multiple chunks. 
The dataset contains 1.2k questions in total, a context length of 128k tokens~(65 chunks), and an average of 4.7 answer changes per question.

\paragraph{\novels}
\novels leverage full-length novels to provide contexts with natural narratives, rich storylines, dynamically interacting characters, and complex, concurrent plotlines. 
For each of the 39 novels included, we generated an initial set of question candidates using Gemini 2.5 Pro~\citep{team2025gemini}, which were subsequently rigorously curated by human experts to ensure quality and consistency. 
We recruited experienced freelancers familiar with the books, investing a total of \$17.4k to validate answers and supporting evidence at each interval. These annotators also refined or replaced low-quality questions, generated missing answer options, and removed ambiguous answer options, resulting in a final dataset filtered down to 55\% of the initial question pool.
The final dataset contains 870 multiple-choice questions (avg. 5.5 options) in total, with an average book length of 150.6k tokens~(77.6 chunks, ranging from 26 to 286), and an average of 4.7 answer changes per question.

\subsection{Evaluation Setup and Metric} \label{sec3: metric}

\paragraph{Setup}
As illustrated in Figure~\ref{fig:overall}, we evaluate models over a sequence of time-interval chunks. 
At each interval, the model is asked the same set of questions based on all the context observed so far, testing its ability to update, incorporate, or retain knowledge on the fly as new information arrives. 
Formally, at interval $t$, a model $M$ observes the set of chunks up to $t$, $\mathcal{S}_t = \{c_i\}_{i=1}^{t}$, and for each question $q_j \in Q$ predicts an answer $p_{j,t} = M(q_j, \mathcal{S}_t)$.\footnote{Our dataset includes sentence-level evidence annotation for each answer, but this would be computationally heavy setup; we therefore use 2k token chunks as the evaluation unit.}

\paragraph{Metric}
We evaluate model performance using interval-level accuracy. At each interval, we compare the model's prediction with the current ground-truth answer, assigning a score of 1 for a match and 0 otherwise. These scores are averaged across intervals to obtain a question-level accuracy, which is subsequently averaged over the entire dataset to compute the final benchmark score.
Evaluation criteria and formal definition of metric in Appendix~\ref{app:eval setups and metric}.

\section{Experimental Setup}
In this section, we share details of the models, context representation, and inference setup. Additional details in Appendix~\ref{app:exp_setup}.
\paragraph{Base Models}
We evaluate 14 LLMs across open-source and proprietary families and a wide range of scales.
We primarily use the Qwen family~\citep{yang2025qwen3, Yang2024Qwen25TR} as the baseline to utilize its multiple sizes (Qwen3-(4B, 8B, 235B), Qwen3-Next-80B, Qwen3-30B~(+Thinking), and Qwen2.5-7B), and include comparable open-source models such as GPT-OSS~\citep{agarwal2025gpt}~(20B, 120B), and Gemma 3~\citep{team2025gemma}~(4B, 27B).
For proprietary models, we focus on Gemini 2.5 (Flash, Pro)~\citep{team2025gemini}, and Gemini 3~\citep{gemini3}.
Unless noted, all models but Gemini are non-thinking versions. 

\paragraph{Context Representations (Base, RAG, Agentic Memory Systems)}
For a \textit{Base} setting, we concatenate all preceding chunks up to the current interval $t$, truncating older chunks when the model's context limit is exceeded.\footnote{For Qwen3-235B, we cap the context to 131k tokens due to GPU memory constraints despite its nominal 1M token capacity.}
For \textit{RAG}, we use Qwen3-Embedding-0.6B~\citep{qwen3embedding} as a retrieval model.
Retrieval is restricted to chunks from previous time intervals. 
Unless otherwise specified, we retrieve the top 30 most relevant memory chunks.
For \textit{agentic memory systems}, we evaluate HippoRAG-V2~\citep{gutierrez2025rag}, MemAgent~\citep{yu2025memagent}, and A-Mem~\citep{xu2025mem}, which maintain and update memory incrementally up to the current time interval.

\paragraph{Inference Setup}
Experiments were mostly conducted with 8 A100 80G GPUs using vLLM~\citep{kwon2023efficient}. 
We use the same setup of temperature 0.7, TopP 0.8, TopK 20, and MinP 0, if not stated otherwise, as the best practice.

\begin{table*}[t!]
    \centering
    \fontsize{8}{10} \selectfont
    \setlength{\tabcolsep}{3pt} 
        \begin{tabular}{lcccccccccccccccccc}
        \toprule
        &&& \multicolumn{8}{c}{\Babi} & \multicolumn{8}{c}{\Novels}  \\
        \cmidrule(lr){4-11} \cmidrule(lr){12-19}
        &&& \multicolumn{4}{c}{Base} & \multicolumn{4}{c}{RAG} & \multicolumn{4}{c}{Base} & \multicolumn{4}{c}{RAG} \\
        \midrule
        Model & Size & Active & \cellcolor{tablegray}All & Sprs. & Mod. & Freq. & \cellcolor{tablegray}All & Sprs. & Mod. & Freq.  & \cellcolor{tablegray}All & Sprs. & Mod. & Freq. & \cellcolor{tablegray}All & Sprs. & Mod. & Freq.  \\
        \midrule
        \multirow{5}{*}{Qwen3}

        & 4B & - &\cellcolor{tablegray}26.4 & 34.4 & 23.1 & 18.5 & \cellcolor{tablegray}29.3 & 37.2 & 25.9 & 21.6 & \cellcolor{tablegray}51.3 & 59.1 & 49.7 & 47.9 & \cellcolor{tablegray}55.8 & 68.1 & 56.0 & 48.3\\
        & 8B &  - &\cellcolor{tablegray}33.1 & 36.8 & 34.0 & 25.8 & \cellcolor{tablegray}35.1 & 39.9 & 34.5 & 28.4 & \cellcolor{tablegray}52.7 & 62.5 & 51.7 & 47.6 & \cellcolor{tablegray}47.7 & 60.0 & 47.8 & 40.4\\
        & 30B & 3B &\cellcolor{tablegray}35.8 & 38.5 & 37.3 & 29.4 & \cellcolor{tablegray}37.8 & 40.1 & 39.2 & 32.0 & \cellcolor{tablegray}62.8 & 71.2 & 63.7 & 57.1 & \cellcolor{tablegray}61.0 & 68.9 & 63.0 & 54.9\\
        & 80B & 3B &\cellcolor{tablegray}41.1 & 44.7 & 42.0 & 34.1 & \cellcolor{tablegray}43.5 & 47.4 & 44.2 & 36.2 & \cellcolor{tablegray}64.6 & 74.5 & 63.3 & 59.9 & \cellcolor{tablegray}65.3 & 74.4 & 65.3 & 59.9\\
        & 235B & 22B &\cellcolor{tablegray}46.8 & 48.4 & 49.8 & 40.0 & \cellcolor{tablegray}44.7 & 46.4 & 46.8 & 39.0 & \cellcolor{tablegray}64.7 & 73.4 & 64.9 & 59.4 & \cellcolor{tablegray}66.0 & 74.7 & 66.5 & 60.6\\
    
        \midrule
        Qwen2.5 & 7B & - &\cellcolor{tablegray}24.7 & 27.3 & 23.6 & 22.4 & \cellcolor{tablegray}32.4 & 35.5 & 32.5 & 27.5 & \cellcolor{tablegray}33.9 & 41.2 & 32.9 & 30.6 & \cellcolor{tablegray}40.2 & 49.6 & 41.0 & 34.2\\
        \midrule
        \multirow{2}{*}{GPT-OSS} 
        & 20B & 3.6B &\cellcolor{tablegray}22.4 & 21.9 & 23.8 & 21.3 & \cellcolor{tablegray}22.2 & 22.2 & 22.9 & 21.2 & \cellcolor{tablegray}45.3 & 54.3 & 45.4 & 39.9 & \cellcolor{tablegray}44.2 & 54.7 & 44.3 & 38.1\\
        & 120B & 5.1B &\cellcolor{tablegray}37.5 & 40.5 & 38.4 & 31.6 & \cellcolor{tablegray}33.7 & 35.9 & 33.9 & 29.8 & \cellcolor{tablegray}54.6 & 62.8 & 53.6 & 50.6 & \cellcolor{tablegray}54.6 & 65.4 & 54.3 & 48.6\\
        \midrule
        \multirow{2}{*}{Gemma 3} 
        & 4B & - &\cellcolor{tablegray}24.2 & 22.9 & 26.3 & 23.3 & \cellcolor{tablegray}25.2 & 24.9 & 26.5 & 23.7 & \cellcolor{tablegray}38.0 & 41.8 & 40.1 & 34.1 & \cellcolor{tablegray}35.2 & 39.4 & 38.1 & 30.5\\
        & 27B & - &\cellcolor{tablegray}37.8 & 40.2 & 40.0 & 30.9 & \cellcolor{tablegray}38.6 & 42.8 & 39.3 & 30.9 & \cellcolor{tablegray}61.2 & 70.1 & 60.5 & 56.6 & \cellcolor{tablegray}56.0 & 67.3 & 57.5 & 48.3\\
        \midrule
        \multirow{2}{*}{Gemini 2.5}      
        & Flash & - &\cellcolor{tablegray}56.2 & 58.3 & 58.6 & 49.3 & \cellcolor{tablegray}56.6 & 60.3 & 58.4 & 47.9 & \cellcolor{tablegray}65.6 & 75.2 & 65.6 & 60.0 & \cellcolor{tablegray}64.5 & 74.2 & 64.3 & 59.0\\
        & Pro & - &\cellcolor{tablegray}60.3 & 64.6 & 62.1 & 50.9 & \cellcolor{tablegray}\textbf{62.2} & \textbf{67.2} & \textbf{63.3} & \textbf{52.9} & \cellcolor{tablegray}\textbf{76.7} & \textbf{83.7} & \textbf{76.4} & 72.9 & \cellcolor{tablegray}\textbf{75.8} & \textbf{82.2} & \textbf{75.4} & \textbf{72.4}\\
        \midrule
        Gemini 3 
        & Pro & - &\cellcolor{tablegray}\textbf{66.3} & \textbf{70.3} & \textbf{69.3} & \textbf{55.6} & \cellcolor{tablegray}51.3 & 51.0 & 55.5 & 45.8 & \cellcolor{tablegray}75.5 & 80.3 & 75.3 & \textbf{73.0} & \cellcolor{tablegray}74.0 & 79.9 & 74.0 & 70.6\\
    
        \bottomrule
        \end{tabular}
    \caption{
    Accuracy~(\%) of models on \babi and \novels under Base and RAG settings. 
    Gemini results are reported with the thinking mode enabled. 
    \textit{Active} denotes the number of active parameters in Mixture-of-Experts~(MoE) models. 
    \textit{Sprs.}, \textit{Mod.}, and \textit{Freq.} denote Sparse, Moderate, and Frequent transition subsets; \textit{All} is averaged over all subsets.
    The highest score within each column is in \textbf{bold}.
    }
    \label{table: overall}
\end{table*}

\section{Evaluation Result}

\subsection{Overall results}
\label{sec5:Overall}
In this section, we analyze key findings from Table~\ref{table: overall},  summarizing the overall performance of the evaluated models on \babi and \novels.

\paragraph{\ours is difficult across various models}
The results indicate that \ours task remains challenging, with substantial room for improvement across all evaluated systems. 
The average accuracy is 33.0\% on \babi and 52.9\% on \novels for open-sourced models; 60.9\% on \babi and 72.6\% on \novels for closed-sourced models.
Even for the strongest closed-source model, Gemini 3 Pro, performance reaches only 66.3\% on \babi and 75.5\% on \novels.

\paragraph{Performance tends to improve with a larger, better base model}
Overall, performance generally scales with model size within the same model family.
Among models of comparable size, the Qwen3 family achieves higher accuracy than alternatives.
Further, comparing Qwen2.5 and Qwen3 shows that a stronger base model consistently leads to better performance.
Proprietary models like Gemini also tend to consistently exhibit higher performance than open-source models.
MoE models perform comparably to fully dense models of similar scale, even with a small number of active parameters~(e.g., Qwen3 30B vs. Gemma 3 27B).

\paragraph{Questions with many transitions are difficult}
When comparing \textit{Sparse}, \textit{Moderate}, and \textit{Frequent} subsets, which are defined by the number of answer changes for each question, we observe that the \textit{Frequent} set, which contains questions with many transitions, poses greater challenges for timely updates. 
On \babi, performance decreases from 42.2\% for \textit{Sparse} questions to 40.6\% for \textit{Moderate} and 33.3\% for \textit{Frequent}; on \novels, the corresponding values are 65.4\%, 57.2\%, and 53.0\% on average.
We hypothesize that this degradation arises because frequent answer changes lead to more dynamic knowledge states, requiring models to repeatedly update multiple facts while retaining previously valid information, which exacerbates both tracking and retention difficulties.

\paragraph{Naive RAG shows limited effectiveness in \ours}
When comparing RAG with the Base setting, we observe different patterns across the models and datasets, but modest drop in performance on average: an improvement of 0.4\% on \textit{Sparse} but a performance drop of 0.04\% on \textit{Moderate} and 0.8\% on \textit{Frequent} subsets.
This suggests that a simple RAG approach is insufficient for \ours, particularly in the \textit{Frequent} subset, where knowledge evolves dynamically, and multiple interrelated facts are distributed across overlapping context chunks. 
We identify two main challenges: (1) retrieval itself becomes difficult when many semantically related chunks exist and questions often require reasoning over multiple chunks~\citep{press2023measuring, shao2025reasonir}; and (2) even with successful retrieval, prior work shows that models are sensitive to input context and often struggle to effectively process complex or irrelevant contexts, which can rather degrade performance~\citep{lee2025corg, mallen2023not}.
Thus, we further analyzed more advanced context representation strategies in Section~\ref{result: agentic_memory}.
Retrieval performance in Appendix~\ref{app:perf_rag_rw}.

\subsection{Thinking improves performance, especially on complex reasoning questions}

\begin{table}[t!]
    \centering
    \fontsize{8}{10}  
    \selectfont
    \setlength{\tabcolsep}{2pt}
        \begin{tabular}{lcccccc}
        \toprule
        Model & Think & \cellcolor{tablegray}All & Tracking & Counting & Bridge & Comparison \\
        \midrule
        \multirow{2}{*}{\makecell[l]{Qwen3-30B}} 
        & - & \cellcolor{tablegray}35.8 & 27.1 & 34.8 & 24.6 & 48.3\\
    & \cmark  & \cellcolor{tablegray}\textbf{43.6} & \textbf{43.8} & \textbf{37.7} & \textbf{37.3} & \textbf{53.9}\\
        \midrule
        \multirow{2}{*}{\makecell[l]{Gemini 2.5\\Flash} } 
    & - & \cellcolor{tablegray}43.2 & 54.4 & 42.6 & 30.0 & 53.1\\
    & \cmark  & \cellcolor{tablegray}\textbf{56.2} & \textbf{58.7} & \textbf{53.9} & \textbf{42.4} & \textbf{69.6}\\
        \midrule
        \multirow{2}{*}{\makecell[l]{Gemini 2.5\\Pro} } 
        & - & \cellcolor{tablegray}42.9 & 52.4 & 45.8 & 29.8 & 50.2\\
    & \cmark  & \cellcolor{tablegray}\textbf{60.3} & \textbf{55.3} & \textbf{57.5} & \textbf{51.0} & \textbf{71.7}\\
        \bottomrule
        \end{tabular}
    \caption{
    Accuracy (\%) of Qwen3-30B and Gemini 2.5 on \babi by question type with and without thinking mode (Think). Best in \textbf{bold}.
    } 
    \label{table: reasoning and thinking}
\end{table}

Inference-time scaling with additional intermediate \textit{thinking} processes~\citep{wei2022chain, openai2024openaio1card, kojima2022large, deepseekai2025deepseekr1incentivizingreasoningcapability} consistently improves performance on \ours. Table~\ref{table: reasoning and thinking} compares models with and without \textit{thinking} mode for Qwen3-30B and Gemini 2.5.
In \babi, most question types other than tracking type require reasoning over multiple pieces of evidence distributed across non-contiguous chunks. Enabling thinking mode leads to consistent improvements in overall accuracy, where the most pronounced improvements are observed in bridge-type questions~(15.4\%). Bridge questions are inherently more challenging as they necessitate multi-hop reasoning, integrating multiple factual sentences while simultaneously tracking various states within the context. In contrast, counting-type questions require tracking only a single state with high precision; thus, the performance gain is relatively marginal~(8.0\%). These results suggest that the internal reasoning process of the thinking mode offers the most substantial benefit when task complexity is high, particularly requiring the simultaneous tracking of multiple independent states.

\subsection{\ours is difficult for even agentic memory systems}

\begin{table}[t!]
    \centering
    \fontsize{8}{10} \selectfont
        \begin{tabular}{lcccc}
        \toprule
        Strategy & \cellcolor{tablegray}All & Sprs. & Mod. & Freq. \\
        \midrule
        Base & \cellcolor{tablegray}24.7 & 27.3 & 23.6 & 22.4\\
        RAG~(30) & \cellcolor{tablegray}\textbf{32.4} & \textbf{35.5} & 32.5 & 27.5\\
        RW~(30) &\cellcolor{tablegray}27.6 & 29.7 & 26.9 & 25.5\\
        RAG~(15) + R.W~(15) & \cellcolor{tablegray}31.5 & 34.0 & 31.8 & 27.2\\
        \midrule
        HippoRAG2 & \cellcolor{tablegray}20.8 & 19.5 & 23.9 & 18.5\\
        MemAgent & \cellcolor{tablegray}31.3 & 30.7 & \textbf{33.6} & \textbf{29.1}\\
        A-Mem & \cellcolor{tablegray}30.3 & 30.6 & 33.3 & 25.6\\
        \bottomrule
        \end{tabular}
    \caption
         {
         Accuracy~(\%) of agentic memory system using Qwen2.5-7B instruct as base model on \babi. 
         The number in parentheses represents the number chunks prepended by each method. Best in \textbf{bold}.
         } 
    \label{table: memory agents}
\end{table}

\label{result: agentic_memory}

Table~\ref{table: memory agents} shows the performance of agentic memory systems built on Qwen2.5-7B-Instruct\footnote{We chose Qwen2.5-7B-Instruct as MemAgent was trained on top of it.} in \babi.
Overall, agentic memory methods underperform naive RAG in aggregated accuracy.
However, on the \textit{Moderate} and \textit{Frequent} subsets, they show competitive or improved performance, with MemAgent achieving the best results.
This is likely due to MemAgent's interval-based memory tracking training objective, which aligns with our evaluation setting that requires continual tracking of dynamic knowledge. 
Nonetheless, its performance remains limited because training is based on static question types with rewards computed only after processing all chunks, rather than at each interval. 
These results highlight \ours as a challenging benchmark even for such agentic memory systems due to its frequent, fine-grained knowledge updates. More analysis in Appendix~\ref{app:rag-rw}.

\section{Analysis}
\label{sec6:analysis}
In this section, we present a detailed analysis of models' behavior on \ours.
We focus primarily on \babi to control for models' prior knowledge~(Appendix~\ref{app:novel-vs-babi}) with Qwen3-30B unless otherwise specified.
Appendix~\ref{app:Analysis} provides additional analyses on \novels, models’ evidence reasoning, and the correlation between \ours and long-context understanding ability.

\begin{table*}[t!]
    \centering
    \fontsize{8}{10} \selectfont
    \setlength{\tabcolsep}{3pt} 
        \begin{tabular}{lcccccccccc}
        \toprule
         &&&  \multicolumn{4}{c}{\text{GT Phase Transitions} (\underline{\textbf{C}}hange)} & \multicolumn{4}{c}{\text{No GT Transition} (\underline{\textbf{S}}tay)}\\
        \cmidrule(lr){4-7} \cmidrule(lr){8-11} 
        \multirow{2}{*}{Models} & \multirow{2}{*}{Size} & \multirow{2}{*}{Think}  & \multirow{2}{*}{\begin{tabular}[c]{@{}c@{}} \texttt{Adaptability} \\ (\textbf{C} / \greencmark) \end{tabular}}  
        & \multirow{2}{*}{\begin{tabular}[c]{@{}c@{}} \texttt{Maladaptation} \\ (\textbf{C} / \redxmark) \end{tabular}}
         & \multirow{2}{*}{\begin{tabular}[c]{@{}c@{}} \texttt{Prescience} \\ (\textbf{S} / \greencmark) \end{tabular}}  
         & \multirow{2}{*}{\begin{tabular}[c]{@{}c@{}} \texttt{Stubbornness} \\ (\textbf{S} / \redxmark) \end{tabular}}    
         & \multirow{2}{*}{\begin{tabular}[c]{@{}c@{}} \texttt{Lag} \\ (\textbf{C} / \greencmark) \end{tabular}}  
        & \multirow{2}{*}{\begin{tabular}[c]{@{}c@{}} \texttt{Volatility} \\ (\textbf{C} / \redxmark) \end{tabular}}
         & \multirow{2}{*}{\begin{tabular}[c]{@{}c@{}} \texttt{Stability} \\ (\textbf{S} / \greencmark) \end{tabular}}  
         & \multirow{2}{*}{\begin{tabular}[c]{@{}c@{}} \texttt{Obstinacy} \\ (\textbf{S} / \redxmark) \end{tabular}}  \\
        &  & && \\
        \midrule
        Gemma 3 & 27B & -  & 31.6 & 28.6 & 11.9 & 27.9 & 12.2 & 27.0 & 25.2 & 35.6\\
        \midrule
        GPT-OSS & 120B   & - & 39.1 & 40.8 & 5.5 & 14.6 & 12.8 & 46.5 & 24.2 & 16.5  \\
        \midrule
        \multirow{2}{*}{Qwen3} & \multirow{2}{*}{30B}
        & -  & 34.3 & 33.6 & 9.7 & 22.4 & 8.9 & 36.7 & 26.3 & 28.1\\
        & & \cmark   & 39.6 & 34.6 & 7.7 & 18.2 & 13.0 & 37.7 & 30.4 & 18.9\\
        \midrule
        \multirow{2}{*}{Gemini 2.5} & \multirow{2}{*}{Flash}
          & -  & 36.3 & 17.0 & 17.2 & 29.5 & 7.5 & 16.9 & 35.0 & 40.7\\
        &  & \cmark & 47.5 & 23.8 & 15.0 & 13.7 & 12.4 & 27.5 & 43.3 & 16.8 \\
        \midrule
        \multicolumn{3}{c}{\textit{Avg}} &  38.1 & 29.7 & 11.2 & 21.1 & 11.1 & 32.1 & 30.7 & 26.1 \\
        \bottomrule
        \end{tabular}
    \caption{
    Analysis of knowledge tracking behavior on \babi. 
    The table shows the average occurrence rate of specific tracking behaviors across all time intervals.
    The second row shows the behavioral types (e.g., \texttt{Maladaptation}) along with the model's predicted action (whether predicted answer \textbf{C}hange or \textbf{S}tay from previous prediction) and the resulting prediction correctness (correct: \textbf{\greencmark} vs. incorrect: \textbf{\redxmark}).
    }
    \label{table: behavior}
\end{table*}

\subsection{Fine-grained analysis of predicted knowledge transition behavior}
\label{sec:behavior analysis}

To analyze how models track evolving factual knowledge, we categorize their behavior under two ground truth (GT) scenarios: \textit{GT Phase Transitions} and \textit{No GT Transition}, where the phase changes or stays, respectively.\footnote{Since Stay intervals are prevalent~(94\% of all intervals), rates are averaged within each GT scenarios to sum to 100\%.}
Table~\ref{table: behavior} shows the average frequency of different behavioral patterns across all intervals. 
The model's predicted transition behavior is defined by two factors: 
(1) whether the model's predicted answer \textbf{C}hanges or \textbf{S}tays relative to the previous prediction, and
(2) whether the resulting answer is correct (\textbf{\greencmark}) or incorrect (\textbf{\redxmark}). 
For descriptive purposes, we map these combinations to specific behavioral archetypes
(e.g., \texttt{Adaptability}). 
We provide a schematic illustration of these behaviors, full results, and additional analysis in Appendix~\ref{app:behavior}.

\paragraph{Enabling explicit thinking alters transition behavior}
Across both Qwen3 and Gemini 2.5, enabling ``thinking mode'' consistently shifts model behavior.
Non-thinking variants exhibit a higher rate of \texttt{Prescience}, \texttt{Stubbornness}, and \texttt{Obstinacy}, indicating that it tends to retain previous answers even when an update is required.
In contrast, thinking-enabled models show a higher rate of \texttt{Adaptability} and \texttt{Stability}, suggesting explicit reasoning improves both transition timing and answer correctness.

\paragraph{Different models exhibit distinct transition behavior}
Comparison over different non-thinking models shows a distinct behavior in how often they change predictions.
GPT-OSS and Qwen3 exhibit a higher overall change rate (\textbf{C}, avg 63.2\%) than stay rate (\textbf{S}, avg. 36.8\%). 
In contrast, Gemini 2.5 and Gemma 3 exhibit a higher stay rate (\textbf{S}, avg. 55.8\%) compared to the change rate (\textbf{C}, avg 44.2\%). 
This is also reflected in their dominant error modes: \texttt{Volatility} and \texttt{Obstinacy}, respectively.
These results suggest model-specific biases toward either \textit{over-updating} or \textit{under-updating} knowledge.

\paragraph{Models often detect true transitions but frequently make unnecessary updates}
We analyzed which behavior is dominant under each GT scenarios.
Under \textit{GT Phase Transitions}, \texttt{Adaptability} is the most frequent behavior (38.1\%), indicating that models often detect true transitions.
In contrast, \textit{No GT Transition} is dominated by \texttt{Volatility}~(32.1\%), showing that models frequently update even when the underlying fact remains unchanged. 
This suggests that models tend to be sensitive to true knowledge updates but also prone to unnecessary changes, likely due to interference from surrounding contextual information.

\subsection{Intra-phase analysis shows trade-off between phase capture and stability}

\begin{table}[t!]
    \centering
    \fontsize{8}{10} \selectfont
    \setlength{\tabcolsep}{5pt} 
        \begin{tabular}{lcccccc}
        \toprule
        Model & Size & Think & ACC & AL($\downarrow$) & DS($\downarrow$) & PM($\downarrow$)  \\ 
        \midrule
        Gemma 3 & 27B & -   & 37.8 & 5.4 & \textbf{26.5}& 30.3\\
        \midrule
        GPT-OSS & 120B   & - & 37.5 & 8.6 & 38.8 & 15.1\\
        \midrule
        \multirow{2}{*}{Qwen3} & \multirow{2}{*}{30B}
        & -   & 35.8 & 6.9 & 34.0 & 23.2\\
        & & \cmark  & 43.6 & 9.0 & 37.0 & 10.4 \\
        \midrule
        \multirow{2}{*}{Gemini 2.5} & \multirow{2}{*}{Flash}
          & -  & 43.2 & 6.2 & 28.3 & 22.3\\
        &  & \cmark  & \textbf{56.2} & \textbf{5.1} & 31.8&  \textbf{7.0} \\
        \bottomrule
        \end{tabular}
    \caption{
    Analysis of intra-phase behavior~(ACC, AL, DS, PM) on \babi over models. 
    Best in \textbf{bold}.
    }
    \label{tab:aldspm}
\end{table}

To evaluate the models' \textit{intra-phase} behavior at finer granularity, we categorize error intervals into three cases and score separately: Acquisition Latency~(AL), Distraction Susceptibility~(DS), and Phase Miss rate~(PM). AL quantifies the delay between a ground-truth state transition and the first correct prediction; DS measures the frequency of incorrect predictions after an initial correct prediction within a phase; PM denotes phases in which the model gets all wrong. 
Same as Accuracy, all metrics are normalized by the total number of intervals per question and averaged across questions.

Table~\ref{tab:aldspm} presents a comparative analysis of metrics across representative models with similar accuracy. 
Although Gemma 3 and GPT-OSS achieve comparable performance, Gemma 3 exhibits a substantially higher PM rate, indicating failure to capture some phase entirely, whereas GPT-OSS, with high \texttt{volatility}, successfully captures each phase at least once, resulting in lower PM but higher AL and DS. 
Comparing the thinking vs. non-thinking model of Qwen3-30B and Gemini 2.5 Flash, we find that accuracy gains are primarily driven by the models' improved ability to capture ground-truth states at least once per phase, as reflected by substantially reduced PM rate. 
However, higher DS of the thinking model implies that contextual distraction, where the model loses track of a previously identified state as the input length increases, remains a persistent challenge.

\subsection{Failure modes vary across question types in \babi}

\begin{figure}[t!]
    \centering
    \begin{minipage}[b]{\linewidth}
        \centering
            \centering
            \includegraphics[width=\linewidth]{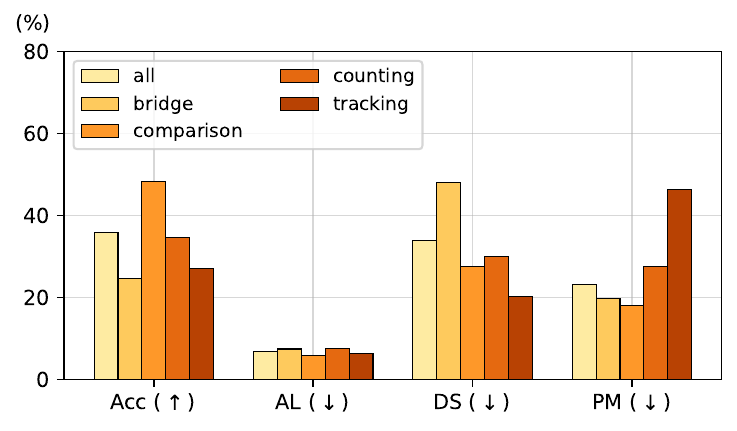}
        \caption{Accuracy~(\%) across different question types of \babi.}% of Qwen3-30B. }
        \label{fig:perf-by-qtype}
    \end{minipage}
\end{figure}

Figure~\ref{fig:perf-by-qtype} analyzes model performance across \babi question types with varying reasoning demands and number of chunks to attend to.
Bridge questions require simultaneous tracking and updating of multiple states, resulting in the highest DS due to increased interference from monitoring multiple entries.
Comparison questions remain relatively high performance, likely not because they are easier, but because the candidate answers are embedded inside the question, similar to multiple choice questions, reducing the search space.
Tracking-type questions, even though involving only a single piece of evidence, remain challenging with the highest PM rate, due to its frequent state transitions~($8.8$ vs. $3.7$-$5.7$ in other types). 
Collectively, these results indicate that different question types induce distinct failure modes in streaming contexts, many of which stem from difficulties in fine-grained fact tracking.

\subsection{Accuracy degrades at later intervals}
\label{sec6:acc-context-length}

\begin{figure}[t!]
    \centering
    \begin{minipage}[b]{\linewidth}
        \centering
            \centering
            \includegraphics[width=\linewidth]{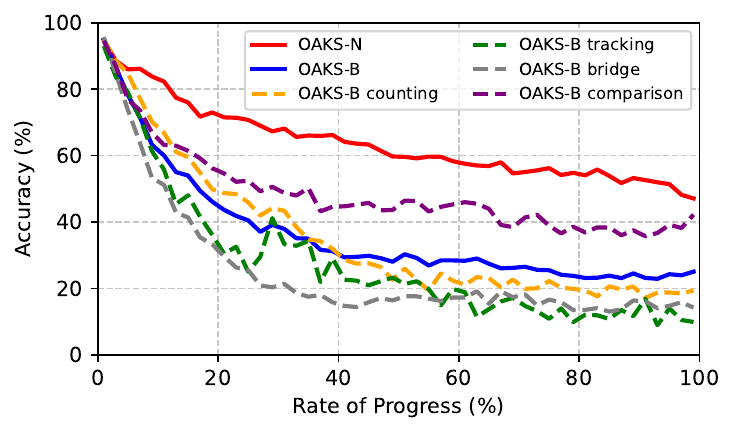}
        \caption{Accuracy~(\%) across the timestep where the question is asked.} 
        \label{fig:perf-by-context-length}
    \end{minipage}
\end{figure}

Figure~\ref{fig:perf-by-context-length} shows that average accuracy at each interval degrades at later intervals.
This degradation is more pronounced in \babi, where the supporting evidence typically appears only once; if the model fails to capture it when it first appears, the error persists and accumulates in subsequent intervals.
Within \babi, bridge and tracking questions exhibit the largest degradation, as they are more vulnerable to missed evidence and compounded errors over time.
In contrast, \novels show more stable performance, likely because relevant information is often revisited across intervals, partially mitigating error accumulation. 

\section{Conclusion}

In this paper, we introduce \Ours~(\ours), a benchmark for evaluating language models in the setting of online adaptation to streaming, continually evolving knowledge. 
We present two datasets, \Babi and \Novels, which contain context incrementally revealed in small chunks over long sequences and fixed questions each with answers annotated for every interval to track evolving facts.
Experiments on 14 models with varying inference strategies show that the task remains challenging even for state-of-the-art models and agentic memory systems, especially under frequent knowledge updates and in later intervals. 
We further find that models are easily distracted by surrounding contextual information and often lose track of previously identified states during the adaptation.

\newpage
\section*{Limitations}

Evaluation on \ours entails extensive inference due to the incrementally accumulating nature of the context. While we have included 14 representative models in this study, due to limited computational costs and API expenses, we could not further test on broader range of architectures. Future work could extend this benchmark to a more diverse set of models to further generalize our findings. 

\noindent Exploring even more complex natural texts with more frequent or complex fact transitions would further enable analysis of scalability and error accumulation.
Also, datasets with contexts that are free from the model's parametric knowledge priors would provide a clearer lens on online adaptation behavior. 

\noindent Our current analysis focuses on inference-time adaptation via incremental context accumulation. However, \ours also serves as a valuable testbed for parametric online learning, and future work could explore how models update their internal weights to assimilate evolving knowledge.

\bibliography{custom}

\newpage
\appendix

\section{Extended Related Work}
\label{app:related_works}
\paragraph{Continual Knowledge Learning and State Tracking Benchmarks}
Real-world knowledge is inherently dynamic: facts can become outdated, remain invariant, or require the incorporation of entirely new information.
Thus, there is growing interest in evaluating language models under settings where knowledge evolves continually~\citep{kim2024carpe, jang2022towards}. 
However, existing benchmarks typically involve a limited number of knowledge updates and often expand the knowledge with divergent facts rather than repeatedly updating the same underlying fact. 
This makes it difficult to isolate and assess a model’s ability to track continual updates to identical pieces of knowledge over time. 
Our work addresses this gap by providing a fine-grained evaluation of dynamic knowledge transitions at the individual fact for the same question set. 

We also draw on research in state tracking, which studies how models maintain evolving states~\citep{kim2023entity, niu2024enhancing}. It focuses on temporal consistency and stepwise updates, which are similar to \ours. However, previous studies typically address short-term, structured states, such as dialogue slots~\citep{budzianowski2020multiwozlargescalemultidomain, lee2019sumbt}, whereas \ours focuses on open-ended, continually updating knowledge states over long horizons in an online setting, evaluating a model's ability to maintain temporal consistency across a long-term continuous stream of information without updating parameters and keeping in context.

\paragraph{Online Adaptation of Streaming Inputs}
In real-world scenarios, knowledge often arrives sequentially over time, requiring models to continuously update their internal knowledge based on streaming inputs, rather than being presented in a static, offline format where all relevant information is available upfront. 
Prior research on online learning in the text domain has focused on self-evolving, lifelong agents that acquire new capabilities or task-level skills over time ~\citep{zheng2025lifelongagentbench, wei2025evo} or synthetic facts with small updates~\citep{wu2023online, lyu2025facttrack}. 
These works emphasize agent competence or behavior, but largely operate at the granularity of \textit{tasks} rather than individual pieces of knowledge. 
However, evaluating fine-grained knowledge updates (i.e., when and how specific facts change) is important, as failures to incorporate small but relevant updates can propagate and accumulate into larger errors when such knowledge is reused or composed in downstream tasks.
Inspired by streaming or online video benchmarks~\citep{lin2024streamingbench, xu2025streamingvlm}, we introduce, to the best of our knowledge, the first benchmark in the text domain that explicitly evaluates streaming knowledge updates at the granularity of individual facts. 

\paragraph{Long-Context Understanding Benchmarks}
With recent LLMs able to process increasingly long contexts, a variety of benchmarks have been proposed to measure long-context understanding. 
Synthetic benchmarks, e.g., Needle-in-a-Haystack~\citep{needle-in-haystack, liu2024lost}, evaluate a model’s ability to retrieve specific information embedded within lengthy inputs, where the target information is sparsely embedded and not strongly correlated with the surrounding context.
Other works adopt more naturalistic or conversational settings. For example, long-context dialogue datasets~\citep{maharana-etal-2024-evaluating, wu2025longmemeval, lee2025realtalk, wan2025storybench} require models to answer questions based on extended conversations or narrative formats. 
There are also benchmarks using long-form documents or novels, where models must answer questions using the full text. 
Moreover, agent-based benchmarks~\citep{hu2025evaluating, wang2025mem} evaluate long-context reasoning through interaction with an external environment. 
Our task is similar in that it also involves long contexts, but differs by focusing on continually updating knowledge in an online setting. 
To capture this, we segment the long context into temporal chunks and annotate answers for each question at every chunk, enabling evaluation of a model’s ability to track and update knowledge over time.

\section{\Ours}

\subsection{Dataset}
\label{app:dataset}

\subsubsection{Data statistics}
\label{sec:data_statistics}

\paragraph{Partition based on frequency of answer changes per question}

\begin{figure}[t!]
    \centering
    \begin{minipage}[b]{\linewidth}
        \centering
            \centering
            \includegraphics[width=\linewidth]{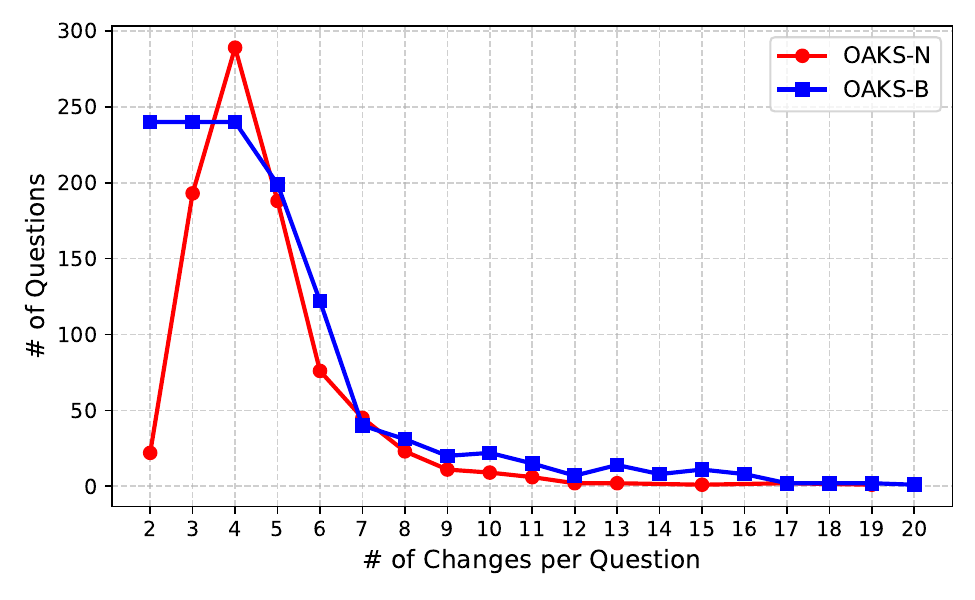}
        \caption{Frequency distribution of answer changes per question for \babi and \novels.}
        \label{fig:data-answer-change-dist}
    \end{minipage}
\end{figure}

We partition each dataset into three subsets based on the frequency of answer changes per question. We define the bins such that the number of samples in each group is approximately balanced, while ensuring that all questions with the same number of changes are assigned to the same group to maintain consistency in our analysis on how the number of answer changes affect on performance. 
The resulting distributions are in Table~\ref{tab:data_stats}. 
Both of our datasets are in English.
Figure~\ref{fig:data-answer-change-dist} shows the frequency distribution of answer changes per question. 
For \babi, the subsets correspond to 2–3, 4–5, and 6–20 changes, respectively; for \novels, the ranges are 2–3, 4, and 5–19 changes.

Since our dataset includes sentence-level evidence annotation for each answer, we could perform sentence-wise fine-grained evaluation using sentence as the data unit. However, this would be computationally expensive, as it would require running inference for every sentence. Therefore, we adopt 2k token chunks as the dataset unit for evaluation.

\begin{table}[t!]
    \centering
    \fontsize{8}{10} \selectfont
    \begin{tabular}{lc|ccc} 
    \toprule
    \textbf{Dataset} & \textbf{Total} & \textbf{Sprs.} & \textbf{Mod.} & \textbf{Freq.} \\ \midrule
    \babi & 1,224 & 40\% & 25\%  & 35\%  \\
    \novels & 870 & 25\% & 33\% & 42\%  \\ 
    \bottomrule
    \end{tabular}
    \caption{Distribution of answer change frequency across datasets: Sparse (Sprs.), Moderate (Mod.), and Frequent (Freq.)}
    \label{tab:data_stats}
\end{table}

\begin{algorithm*}[t]
    \caption{\Babi dataset construction}
    \label{algorithm:babi_dataset_construction}
    \begin{algorithmic}[1]
    \REQUIRE Narrative facts $\mathcal{F}$, Question Templates $\mathcal{P}$, Canonical Verb Groups $\mathcal{V}$
    \ENSURE Dataset $\mathcal{D} = \{(Question, Timeline\_Answers)\}$
    
    \STATE $\mathcal{S} \leftarrow \text{Empty entity state dictionary}$
    \STATE \COMMENT{1. Knowledge Extraction and State Logging}
    \FOR{each $fact_i$ in $\mathcal{F}$}
        \STATE $(s, v, o, [r]) \leftarrow \text{ParseSentence}(fact_i)$ 
        \STATE $v_{canon} \leftarrow \text{Normalize}(v, \mathcal{V})$
        \STATE $\mathcal{S}[s][v_{canon}] \leftarrow \mathcal{S}[s][v_{canon}] \cup \{(o, t)\}$
        \STATE $\mathcal{S}[o][v_{canon}] \leftarrow \mathcal{S}[o][v_{canon}] \cup \{(s, t)\}$
        
        \IF{$v_{canon} = \text{VERB\_P}$ (Transfer)}
            \STATE $\mathcal{S}[r][\text{get}] \leftarrow \mathcal{S}[r][\text{get}] \cup \{(o, s, t)\}$
        \ENDIF
        
    \ENDFOR
    \STATE \COMMENT{2. Temporal Question Synthesis}
    \STATE $\mathcal{D} \leftarrow \emptyset$
    \FOR{each $entity$ in $\mathcal{S}$}
        \FOR{each $template$ in $\mathcal{P}$}
            \STATE $Q \leftarrow \text{FillTemplate}(template, entity)$
            \STATE $Actions \leftarrow \text{RetrieveActions}(\mathcal{S}, entity, template)$
            \STATE $T \leftarrow \text{GenerateTimeline}(Actions, \text{length}=|\mathcal{F}|)$
            \IF{$\text{CountChanges}(T) > 0$} 
                \STATE \text{// Save only the ones with more than one change}
                \STATE $\mathcal{D} \leftarrow \mathcal{D} \cup \{(Q, T)\}$ 
            \ENDIF
        \ENDFOR
    \ENDFOR
    \RETURN $\mathcal{D}$
    \end{algorithmic}
\end{algorithm*}

\begin{table*}[t]
    \centering
    \small
    \begin{tabular}{p{0.3\linewidth}p{0.6\linewidth}}
    \toprule
    \textbf{Category} & \textbf{Question Template} \\ \midrule
    \multirow{3}{*}{\textbf{Tracking Questions}}
     & Where is [SUB]? \\
     & Who is holding [OBJ]? \\
     & Who gave the [OBJ] to someone else? \\ \midrule
    \multirow{14}{*}{\textbf{Counting Questions}}
     & How many times has [SUB] moved to [PLACE]? \\
     & How many times has [SUB] moved? \\
     & How many people have visited [PLACE]? \\ 
     & How many times has [SUB] picked up [OBJ]? \\
     & How many unique objects has [SUB] picked up? \\
     & How many times has [SUB] dropped [OBJ]? \\
     & How many unique objects has [SUB] dropped? \\
     & How many total times has [OBJ] been picked up? \\
     & How many unique people have held [OBJ]? \\
     & How many total times has [OBJ] been dropped? \\
     & How many unique people have dropped [OBJ]? \\ 
     & How many times has [RECIPIENT] received anything from anyone? \\
     & How many different people have given something to [RECIPIENT]? \\
     & How many times has [SUB] given any object to [RECIPIENT]? \\ 
     \midrule
     \multirow{6}{*}{\textbf{Bridge Questions}}
     & Who most recently traveled directly from [PLACE1] to [PLACE2]? \\
     & Where was [SUB1] the last time [SUB2] moved to the [PLACE]? \\
     & Where is the most recent location that [SUB] moved to after acquiring [OBJ]? \\
     & Where is the most recent location that [SUB] moved to after dropping [OBJ]? \\
     & Where is the most recent location that [SUB] acquire [OBJ]? \\
     & Where is the most recent location that [SUB] drop [OBJ]? \\ \midrule
     \multirow{6}{*}{\textbf{Comparison Questions}}
     & Has [SUB1] or [SUB2] visited more distinct places? \\
     & Which location did [SUB] visit more often, [PLACE1] or [PLACE2]? \\
     & Who picked up a greater number of distinct objects, [SUB1] or [SUB2]? \\
     & Who dropped a greater number of distinct objects, [SUB1] or [SUB2]? \\
     & Who picked up a greater number of objects, [SUB1] or [SUB2]? \\
     & Who dropped more objects, [SUB1] or [SUB2]? \\ \bottomrule
    \end{tabular}
    \caption{Question templates used for \babi dataset construction. The slots [SUB], [OBJ], [PLACE], and [RECIPIENT] are dynamically filled based on the entities present in the narrative facts.}
    \label{tab:babi_q_template}
\end{table*}

\subsubsection{\Babi}
\label{app:babi}

\paragraph{BABILong}
\babi uses a sample of context from BABILong~\citep{kuratov2024babilong}, which builds on the bAbI benchmark~\citep{weston2015towards} fact-based world state (e.g., "Mary moves from kitchen to hallway"), which is interleaved within long-form novels to evaluate retrieval and reasoning under high distractor density.
BABILong is under Apache License, Version 2.0, and the license of the dataset used when constructing BABILong is PG-19 corpora~\citep{raecompressive2019} under Apache 2.0 License and bAbI dataset~\citep{weston2015towards} under BSD License.

\paragraph{Dataset Construction Process}

We sampled 12 examples from the original BABILong dataset~\citep{kuratov2024babilong}, each containing a sufficient number of facts (average 87) to generate questions that capture knowledge transitions, with an average of 4.7 answer changes per question (ranging from 2 to 20). 
Algorithm~\ref{algorithm:babi_dataset_construction} illustrates how we construct the four question types, where the question templates $\mathcal{P}$ are listed in Table~\ref{tab:babi_q_template} and the canonical verb groups $\mathcal{V} \in \{\text{Move, Acquire, Discard, Transfer}\}$. 
The algorithm can be easily applied to generate additional questions whenever a sufficient number of transitional facts are available.
To avoid potential confusion between the novel content and original bAbI facts, we modified location and character names. 
Texts are split into chunks of 2k tokens using the GPT-NeoX tokenizer. 

\paragraph{Data Statistics and Question Types of \babi}
Tracking questions focus on a single state but involve frequent fact updates, experiencing 9 state updates per question, while others have 4.6 on average. 
Comparison and bridge questions require reasoning over multiple context chunks, while counting questions require tallying actions or occurrences. 
Representative examples and the distribution of question types are provided in Table~\ref{tab:babilong_examples}.
Overall, the dataset contains a total of 1.2k questions, context split into 65 chunks of 2k tokens and including 87 facts on average, and the average number of answer changes per question is 4.7.

\begin{table*}[t]
    \centering
    \renewcommand{\arraystretch}{1.3}
    
    \fontsize{8}{10} \selectfont
    \begin{tabular}{l c p{5cm} p{5cm}}
    \toprule
    \textbf{Type} & \textbf{Percentage} & \textbf{Description} & \textbf{Example} \\ 
    \midrule
    Tracking & 7\% 
    & Questions requiring only a single piece of evidence. 
    & \textit{Where is Daniel?} \\ 
    \midrule
    Counting & 28\% 
    & Questions requiring multiple pieces of evidence and simple counting. 
    & \textit{How many times has Sandra moved?} \\ 
    \midrule
    Bridge & 30\% 
    & Questions requiring multiple pieces of evidence. 
    & \textit{Who most recently traveled directly from office to kitchen?}\\ 
    \midrule
    Comparison & 35\% 
    & Questions requiring multiple pieces of evidence and simple comparative reasoning. 
    & \textit{Who dropped more objects, Sandra or Mary?} \\ 
    \bottomrule
    \end{tabular}
    \caption{Overview of Question Types, Statistics, and Examples for \Babi dataset. This table describes the four main categories of questions found in the dataset.}
    \label{tab:babilong_examples}
\end{table*}

\subsubsection{\Novels}
\label{app:novels}

\paragraph{Book Selection} 

We utilize novels as the foundational context for generating questions in \Novels, as they present multiple entities whose states evolve dynamically alongside the narrative.
Moreover, literary narratives frequently incorporate flashbacks, temporal jumps, and speculative future scenarios, all of which make it difficult for models to accurately track and integrate temporal dynamics. 
The evidence for each question is subtly woven into the text, mirroring the nuanced, context-rich situations language models face in real-world use.
To ensure a sufficient density of state transitions in a story, we select novels from adventure, mystery, and science-fiction genres, which are plot-driven or narratively rich.

\paragraph{Dataset Construction Process}
\label{app:novels-dataset-construction}

After obtaining the main narrative text\footnote{We removed non-content elements such as titles, author names, and tables of contents, as these could interfere with model evaluation by hinting prior knowledge acquired during pretraining.}, we segmented the text into chunks of approximately 2,000 tokens using the gpt-neox tokenizer. To preserve narrative coherence, sentences separated by newline characters in the original text were kept within the same chunk whenever possible, preventing semantically related content from being split across chunks.

After annotation, each question included between 5 and 15 answer options (5.5 on average), comprising both correct and distracting options. By default, all questions included the option "We cannot answer this question at this point," which was intended to be selected only before the relevant information appears in the narrative. Once a valid answer option becomes available, this option never becomes an answer. The order of answer options was randomized when finalizing the dataset.

Representative examples of the annotated questions and answer transitions are provided in Table~\ref{tab:novel_dataset_examples}.

\begin{table*}[t]
    \centering
    \fontsize{8}{10} \selectfont
    \begin{tabular}{c c c c c c c c c c c c c c c c c c c c c}
    \toprule
    \multicolumn{12}{p{0.45\linewidth}|}{
        \raggedright 
        \textbf{Example Question 1:} \newline What mode of transportation are Phileas Fogg and Passepartout currently utilizing?  \newline \newline \textit{Source: Around the World in Eighty Days}
    } & 
    \multicolumn{9}{p{0.45\linewidth}}{
        \raggedright
        (A) Train from London to Paris \newline (B) The steamer 'Mongolia' \newline (C) Train from Bombay to Calcutta \newline (D) An elephant \newline (E) Train through India \newline (F) The steamer 'Rangoon' to Hong Kong  \newline (G) The question cannot be answered at this point in the story.
    } \\
    \midrule
    
    \textbf{Chunk} & \textbf{1} & \textbf{2} & \textbf{3} & \textbf{4} & \textbf{5} & \textbf{6} & \textbf{7} & \textbf{8} & \textbf{9} & \textbf{10} & \textbf{11} & \textbf{12} & \textbf{13} & \textbf{14} & \textbf{15} & \textbf{16} & \textbf{17} & \textbf{18} & \textbf{19} & \textbf{$\cdots$} \\
    
    Answers & G & A & A & B & B & B & B & C & C & D & D & D & D & E & E & F & F & F & F & $\cdots$\\
    \bottomrule
    \toprule
    \multicolumn{12}{p{0.45\linewidth}|}{
        \raggedright 
        \textbf{Example Question 2:} \newline What is Elizabeth's opinion of Mr. Darcy?  \newline \newline \textit{Source: Pride and Prejudice}
    } & 
    \multicolumn{9}{p{0.45\linewidth}}{
        \raggedright 
        (A) He is proud and disagreeable. \newline (B) She is ashamed of her prejudice and feels respect for him.  \newline (C) She loves him. \newline (D) He is a dishonest man who does not keep his word. \newline (E) She is offended by his perception of her. \newline (F) The question cannot be answered at this point in the story.
    } \\
    \midrule
    
    \textbf{Chunk} & \textbf{1} & \textbf{2} & \textbf{3} & \textbf{4} & \textbf{5} & \textbf{6} & \textbf{7} & \textbf{8} & \textbf{9} & \textbf{10} & \textbf{11} & \textbf{12} & \textbf{13} & \textbf{14} & \textbf{15} & \textbf{16} & \textbf{17} & \textbf{18} & \textbf{19} & \textbf{$\cdots$} \\
    
    Answers & F & F & F & A & A & A & A & A & A & D & D & D & D & E & E & E & A & B & B & $\cdots$\\
    \bottomrule
    \toprule
    \multicolumn{12}{p{0.45\linewidth}|}{
        \raggedright 
        \textbf{Example Question 3:} \newline What is Victor Frankenstein's primary goal or motivation at this point in the story?  \newline \newline \textit{Source: Frankenstein}
    } & 
    \multicolumn{9}{p{0.45\linewidth}}{
        \raggedright 
        (A) To learn the secrets of nature and, specifically, the principle of life. \newline (B) To escape the memory and consequences of his work.  \newline (C) To destroy the Monster as an act of vengeance. \newline (D) Recovering from his illness \newline (E) Living in regret and remorse because his creation murdered his brother William. \newline (F) To create a companion for the Monster. \newline (G) The question cannot be answered at this point in the story.
    } \\
    \midrule
    
    \textbf{Chunk} & \textbf{1} & \textbf{2} & \textbf{3} & \textbf{4} & \textbf{5} & \textbf{6} & \textbf{7} & \textbf{8} & \textbf{9} & \textbf{10} & \textbf{11} & \textbf{12} & \textbf{13} & \textbf{14} & \textbf{15} & \textbf{16} & \textbf{17} & \textbf{18} & \textbf{19} & \textbf{$\cdots$} \\
    
    Answers & G & G & A & A & A & A & D & D & D & D & E & E & E & C & B & B & B & B & B & $\cdots$\\
    \bottomrule
    \end{tabular}
    \caption{\Novels Example Questions, Answer Choices, and Answer Evolution over Book Chunks. \textit{Note: The Answers shown in the diagram are for demonstration purposes only and may not accurately reflect the ground truth answers for the listed chunks.}}
    \label{tab:novel_dataset_examples}
\end{table*}

\paragraph{QA draft generation}
\label{app:qa_draft_generation}

Initial drafts of questions and answer options for each book were generated using Gemini 2.5 Pro~\citep{team2025gemini}. To ensure broad coverage of the narrative and to capture diverse aspects of the story, we adopted a two-step generation procedure. First, we identified the main entities in each novel—such as central characters or key objects—by prompting the model with the full text of the book (see Prompt~\ref{gemini_prompt_character}). Second, for each identified entity, we generated questions that synthesize information distributed across multiple chunks of the narrative. These questions were designed to either require multi-hop reasoning or track states that evolve over the course of the story~(see Prompt~\ref{gemini_prompt_question}).

\begin{figure*}[t]
\begin{promptcolorbox}[label=gemini_prompt_character]{Prompt}{Identifying Main Entity}{
You are an AI assistant specializing in narrative analysis and entity extraction. Your task is to analyze the novel and identify the most important characters and objects, central to the narrative.\\\\
Your response must only contain the names of the entities, separated by a | character. Do not include labels, explanations, or any other text.\\\\
Example Output:\\
Jason Hill | Maria | The Sapphire | Captain Eva | The Northern Village\\
}
\end{promptcolorbox}
\end{figure*}

\begin{figure*}[t]
\begin{promptcolorbox}[label=gemini_prompt_question]{Prompt}{Generating Question about a Specific Entity}{
You are given a novel. Your task is to generate a set of Question, Answer, and Evidence pairs that fulfill the following criteria:\\
\\
\#\#\# Question Criteria\\
- \textbf{Synthesizing Information}: Each question must require synthesizing information from at least \textit{two} separate parts of the novel. A question that can be answered with a single local quote should not be included.\\
- \textbf{Types of questions}: Example topics include:\\
\hspace*{1.5em} - Tracking the evolution of a character's feelings, understanding, or situation over time.\\
\hspace*{1.5em} - Connecting a character's backstory from one chapter to their actions in another.\\
\hspace*{1.5em} - Solving a mystery for which clues are scattered across different locations.\\
\hspace*{1.5em} - Comparison questions where the relevant information appears in different parts (e.g., "Who is older, A or B?").\\
- \textbf{Avoid single-point questions}: Avoid questions that can be answered by quoting just one part of the novel.\\
\\
\#\#\# Answer \& Evidence Format\\
- \textbf{Answer}: Provide a \textit{concise} answer in \textit{a few words}.\\
- \textbf{Evidence}: Then, \textit{support} the answer with quoted pieces of evidence from the text. These evidences should come from different chunks of the novel (not from the same chunk).\\
\\
\#\#\# Input Schema\\
You will receive two inputs:\\
1.  A JSON object: This object contains the Novel text. \\
2.  An Entity: A string representing the main character or object to be tracked (e.g., "Jason Hill").\\
\\
\#\#\# Output Schema\\
The output must be a single JSON object matching this structure. 
\begin{Verbatim}
[{
   "question": "What is Harry's primary Quidditch broomstick?",
   "answers": [{
      "answer": "Nimbus Two Thousand",
      "source": "One of Harry’s most prized possessions was his Nimbus Two 
        Thousand racing broom."
     },
     {
      "answer": "A school-owned Shooting Star",
      "source": "He had been riding one of the school brooms at team practice, 
        an ancient Shooting Star, which was very slow and jerky; he definitely 
        needed a new broom of his own."
     },
     ...
    ]
 },
]
\end{Verbatim}

\#\#\# \textbf{ENTITY TO TRACK}
\begin{Verbatim}
{entity}
\end{Verbatim}
% }
}
\end{promptcolorbox}
\end{figure*}

\paragraph{Reason for choosing Multiple Choice}
While answers in \babi are typically stated explicitly, open-ended generation for \novels often yields many valid surface forms for the same underlying answer. This makes automatic evaluation less reliable~\citep{bai2024longbench2, wang2024novelqa}. To enable consistent scoring, we therefore cast \novels as a multiple-choice QA task. Each question has at least five options, including distractors, to keep the difficulty consistent and to challenge models that have not closely tracked the narrative details.

\paragraph{Manual Curation and Quality Control}
\label{app:novel_quality}

To ensure that each question is objectively answerable at any point in the narrative, annotators filtered or refined low-quality questions, created new questions, aligned answer options with the source text, and annotated explicit supporting evidence for each state transition. Through this rigorous process, only 55\% of the initial questions were retained or re-formulated as high-quality. 

Discarded questions mostly resembled conventional reading-comprehension questions that are answerable only after reading the entire book. The removed questions primarily fell into three categories:
(i) questions that merely stitched together information from multiple chunks thus reduced to evidence searching rather than state tracking;
(ii) questions that exhibited only a single state change but did not require multi-chunk reasoning or multi-hop inference; and
(iii) questions whose answer options were unsupported by the surrounding context or appeared simultaneously within the same chunk.

Beyond filtering out low-quality questions, the annotators and authors extensively revised the remaining questions to ensure that each question was objectively answerable at any point in the narrative and that exactly one answer option was correct at each chunk. For example, questions that asked about a state's evolvement were revised to ask the status at specific point in the story.
(e.g., \textit{ How does Mr. Darcy's assessment of Elizabeth Bennet's appearance evolve?} \( \rightarrow\) \textit{What is Mr. Darcy's latest assessment of Elizabeth Bennet's appearance?} )
Similarly, when a state was revealed cumulatively over time, we reformulated questions to target the most recent revelation(e.g., \textit{What are the different names or identities Erik is known by?} \( \rightarrow\) \textit{What term has been most recently introduced to describe the man who lives at the opera?}), rather than the entire accumulated state, ensuring that exactly one answer option is correct at each point in the narrative.

Answer options were also carefully curated: unsupported options were removed, contextually necessary options were added, and cases where multiple options appeared within the same chunk were consolidated. This process ensured that the answers remain well-defined and that only a single option was valid for each chunk. 
Throughout this process, the validity of the question or answer options and the correctness of the selected answers were jointly reviewed and discussed by at least two annotators or authors.

\paragraph{Human Annotation}
\label{app:human annotation}

Via Upwork, we hired 18 experienced freelancers who were native English speakers from USA and had already read the selected books. Compensation per book ranged from \$50 to \$850, depending on book length and the number of initial draft questions, with a total annotation cost of \$17,400. 
Beyond the core question-answer annotation task, compensated at approximately \$20 per hour with an average workload of 14 hours per book, we additionally provided payment for high-quality question generation and substantial question–answer revisions.

Annotators were provided with (i) the novel text segmented into chunks and (ii) a spreadsheet containing the questions and answer options. For each question, annotators labeled the correct answer at every chunk, and revised or removed questions and answer options.
Whenever an answer changed from previous answer, annotators were required to copy and paste the exact evidence sentence from the corresponding chunk that justified the answer transition.

The authors subsequently conducted a thorough verification of all annotated files to ensure consistency and correctness. This included checking that (i) each question had exactly one valid answer per chunk, (ii) the cited evidence explicitly appeared in the specified chunk and correctly supported the selected answer, (iii) the cited evidence aligns with the chunk the answer was chosen, and (iv) each question contained at least five answer options, including distracting options.

The full annotation instructions provided to the annotators are included in Instruction~\ref{annot_inst} to \ref{annot_inst_gen_options}.
To ensure high-quality annotations and alignment with our research goals, we provided the annotators with an explanation on academic purpose of the study.
The annotation process was conducted transparently through the Upwork platform, ensuring fair labor practices, after internal ethics review.

\begin{figure*}[t]
    \begin{instructioncolorbox}[label=annot_inst]{Instruction}{Overall Instruction to the Annotators}
    \textbf{Task Instruction}: \\
    We have divided the assigned book into multiple chunks. The goal of this project is to `collect and refine questions and answers that reflect “dynamic changes throughout the storyline”.` In other words, tracking the changing answers in each chunk. For example, if the main character moves frequently, a question like ‘where is the main character living?’ will have different answers at different points in the story, depending on the events that have occurred so far.\\
    \\
    \textbf{Your responsibilities}:\\
    You will be working with an Excel file that contains text from a book, along with questions and answer options. Please fill in the Excel file according to the provided guidelines. See [Guidelines on how to fill in Excel] for details.\\
    
    - Tasks:
    \begin{enumerate}[leftmargin=1.5em, nosep, label=\arabic*.]
        \item Read the given chunk carefully.
        \item After reading each chunk:
        \begin{itemize}[leftmargin=1.2em, nosep, label=--]
            \item Answer the given multiple-choice questions based on \textbf{the content in the current chunk and all previous chunks.} If none of the existing options are appropriate, add a new one. (Read [Guidelines for Answering] for more details)
            \item Copy and paste the relevant evidence snippet from the chunk to the `source` tab.
        \end{itemize}
        \item Generate at least five new questions that fit the same dynamic-tracking criteria. (Read [Guidelines for Generating Questions] for more details)
        \item Generate \textbf{distracting} answer options. (Read [Guidelines for Generating Distracting Answer Options] for more details)
    \end{enumerate}
    
    \end{instructioncolorbox}
\end{figure*}

\begin{figure*}[t]
    \begin{instructioncolorbox}[label=annot_inst_answering]{Instruction}{Guidelines for Answering}
    
    \begin{enumerate}[leftmargin=0.5em, nosep, label=\arabic*.]
        \item Use only information from the current and previous chunks. Do NOT use prior knowledge about later parts of the story.
        \item Choose the most accurate answer based on the information available so far. Each question should have only \textbf{one} appropriate answer.
        \begin{itemize}[leftmargin=1.2em, nosep, label=--]
            \item If a question appears to have multiple correct answers, and both seem valid:
            \begin{itemize}[leftmargin=1.2em, nosep, label=$\bullet$]
                \item Write “remove” for the option that has never been chosen as an answer before in Column E. Make sure you do not choose the option later on.
                \item If removing an option affects later chunks, please contact us for clarification.
                \item Examples
                \begin{itemize}[leftmargin=1.2em, nosep, label=$\circ$]
                    \item \textbf{[Example 1]} Question is \textit{"Where is character A living?"} with options "Paris" and "LA". Earlier answer is \textit{Paris} and New information states that \textit{A lives in LA}\\                        
                            \(\rightarrow\) Choose LA as the correct answer and provide evidence.                         
                    \item \textbf{[Example 2]} Question is \textit{"What is the role of A?"} with a option "nurse". In Chunk 5, it mentions A is a \textit{nurse} and in chunk 7, it adds that \textit{A also started working as a writer on weekends}\\
                            \(\rightarrow\) Add a new option \textit{“nurse and writer”} and choose this as an answer.
                \end{itemize}
            \end{itemize}
            \item Along with several other answer options, there are always two default options:
            \begin{itemize}[leftmargin=1.2em, nosep, label=$\bullet$]
                \item "Prev" : Previous answer still holds
                \begin{itemize}[leftmargin=1.2em, nosep, label=$\circ$]
                    \item \textbf{[Example]} If in previous chunks the character moved to home A, and the current chunk doesn’t mention about move, then the answer to “Where is the character living” remains A and you should choose "Previous answer still holds"
                \end{itemize}
                \item "NA" : We cannot answer to this question at this point
                \begin{itemize}[leftmargin=1.2em, nosep, label=$\circ$]
                    \item If the answer cannot be found because the relevant information has not appeared in the story yet, choose “NA.”
                    \item \textbf{[Example]} If the question is “Where is the character living?” and the location has not yet been revealed, the answer should be “NA”.
                    \item Once you select any answer other than "NA", you should not choose "NA" afterwards. If none of the existing options fit, add a new answer option instead.
                \end{itemize}
            \end{itemize}    
        \end{itemize}
        
        \item Except for "NA" or "Prev", copy and paste the relevant evidence snippet from the chunk.
        
        \item If answering a question requires information from multiple chunks, you may select the answer only when all necessary evidence is present in the current or previous chunks. In this case, the evidence should be taken from the latest chunk—the one that contains the final piece of information needed to answer the question.
            
        \item If the current chunk includes a flashback (a reflection on a past location), do NOT treat that as the current state.
            
        \item If all the answer options are inappropriate, create and add new option.
                
        \item When a question asks about a character’s understanding of something, include evidence showing \textit{where the character actually understands} the concept or situation. However, sometimes a character \textit{hears} or \textit{learns} about a situation but becomes confused because it contradicts their prior beliefs. In those cases, only select the answer and evidence that \textbf{explicitly show the character has reached understanding}.
            \begin{itemize}[leftmargin=1.2em, nosep, label=--]
                \item \textbf{Valid example}: “Character A nodded in agreement after the explanation.” (This shows clear understanding.)
                \item  \textbf{Invalid example}: “Character A listened to the explanation.” (This only shows the character hearing the explanation, not understanding it yet.)
            \end{itemize}
        
    \end{enumerate}
    
    \end{instructioncolorbox}
\end{figure*}

\begin{figure*}[t]
    \begin{instructioncolorbox}[label=annot_inst_gen_questions]{Instruction}{Guidelines for Generating Questions}
    Valid question set satisfies these criteria: 
    \begin{enumerate}[leftmargin=0.5em, nosep, label=\arabic*.]
        \item There could be two types of questions:
        \begin{itemize}[leftmargin=1.2em, nosep, label=--]
            \item The question tracks the evolution of a character's feelings, understanding, or situation as the story progresses. The appropriate answer to the question is different according to the relative position in the story. 
            \item A question that you can answer ONLY when you have information from two or more chunks. 
        \end{itemize}
        \item  For each chunk, there should be only one answer for each question.
    \item  Do not ask anything trivial.
        \begin{itemize}[leftmargin=1.2em, nosep, label=--]
        \item For example, what is the color of the character A wearing?
        \end{itemize}
    \item  The question shouldn’t be subjective or ambiguous
        \begin{itemize}[leftmargin=1.2em, nosep, label=--]
        \item For example, “who is character A’s best friend?” since the definition of “best friend” is ambiguous unless explicitly mentioned in the book.
        \end{itemize}
    \item  You have to generate at least 5 new questions over the whole book. Make sure you do the same answer, evidence annotation for these new questions as you did for original questions.
        
    \end{enumerate}
    
    \end{instructioncolorbox}
\end{figure*}

\begin{figure*}[t]
    \begin{instructioncolorbox}[label=annot_inst_gen_options]{Instruction}{Guidelines for Generating Distracting Answer Options}
    \begin{enumerate}[leftmargin=0.5em, nosep, label=\arabic*.]
        \item The distracting options should be plausible but incorrect, designed to challenge readers who haven’t paid close attention to the details of the story.
        \item Each question must have at least six answer choices in total. These two default options (”NA” and "Prev") are always included, and the additional four options should always exist. If the answer options are fewer than six, add the incorrect distracting answers to reach a minimum of six total choices.
    \end{enumerate}
    
    \end{instructioncolorbox}
\end{figure*}

\subsubsection{Dataset Safety}
To ensure data privacy and ethical standards, for \babi, we modified all location and character names using randomized, non-existent strings to ensure they do not refer to real-world individuals or locations. The factual sentences in the widely used original bAbI benchmark~\citep{weston2015towards} consist of mundane descriptions of daily activities and contain no offensive content. For \novels, the foundational contexts are derived from existing, publicly available novels.

\subsection{Evaluation Setups and Metrics}
\label{app:eval setups and metric}

\subsubsection{Evaluation Criteria}
After extracting the final answer from the model’s prediction, we compare it against the ground-truth answer using normalized text. To ensure a robust benchmark, correctness is assessed using an exact-match criterion.
For \novels, where models are prompted to respond with a single character corresponding to an answer option, we extract and evaluate that character directly.

For \babi, to mitigate the ambiguity of open-ended generation, we conditionally allow multiple equivalent answer forms only under the following cases:
(i) Counting-type questions: we accept both numeric and textual representations (e.g., once, twice, three times).
(ii) Counting-type questions before the relevant information is revealed: we accept both Unknown and 0.
(iii) Comparison-type questions before valid evidence is encountered: we accept both Unknown and Same.

\subsubsection{Metrics}
\label{app:metric-equation}

For \Babi and \Novels, we evaluate a model with Accuracy and analyze fine-grained behavior over three additional metrics: Acquisition Latency~(AL), Distraction Susceptibility~(DS), and Phase Miss rate~(PM).

\paragraph{Accuracy}
Accuracy measures the proportion of time intervals where the model predicts the correct answer for a given question. We then average the score over all questions. We provide a formal definition below.

Given our dataset consisting of $C$ chunks $\mathcal{C} = \{c_{i}\}_{i=1}^{C}$ and a question set $\mathcal{Q} = \{q_{j}\}_{j=1}^{Q}$ with corresponding ground truth answers $\mathcal{A} =\{a_{i,j}\}_{j=1, i=1}^{Q, C}$, a model $M$ is asked to predict an answer given a set of chunks $\mathcal{S}_{t} = \{c_{i}\}_{i=1}^t$ up to a certain interval $t$. Let the model's prediction be $p_{j,t} = M(q_{j}, \mathcal{S}_{t})$.

Then the Accuracy of model $M$ is defined as:
\begin{equation}
    \textbf{Accuracy} = \frac{1}{Q}\sum_{j=1}^{Q} \frac{1}{C}\sum_{i=1}^{C} \mathds{1} \left[p_{i,j} = a_{i,j}\right]
\end{equation}

\paragraph{Acquisition Latency~(AL)} 
AL describes how quickly or slowly it takes the model to adapt to a new state relative to the length of all sequence. For each phase, we count the number of incorrect predictions before a correct prediction (\textit{lag} intervals) and obtain the proportion of the whole intervals. Then, we compute the average over all the questions.

Mathematically, 

\begin{enumerate}[leftmargin=1.5em, nosep, label=$\bullet$]
    \item let $N_j$ be the total number of distinct phases for question $q_j$,
    \item let $T_{k,j}$ be the $k$-th phase for $q_j$ and $|T_{k,j}|$ its duration,
    \item let $a_{t,j}$ be the ground truth answer at interval $t$ for question $q_j$.
\end{enumerate}

Define
$$
\tau_{k,j} = \min\{t \in \{1,\dots, |T_{k,j}|\} \; |\; p_{t, j} =  a_{t,j}\}
$$
as the first time step in phase $T_{k,j}$ where the models answers $q_j$ correctly. If the model is never correct, we define $\tau_{k,j} = 0$.

AL metric is defined as:
\begin{equation}
    \textbf{AL} = \frac{1}{Q}\sum_{j=1}^{Q} \frac{1}{C} \sum_{k=1}^{N_j}( \tau_{k,j}-1 )\cdot \mathds{1}[\tau_{k,j} > 0]
\end{equation}

\paragraph{Distraction Susceptibility~(DS)}
DS measures the distraction rate of a model. For each phase, we quantify DS by counting the number of incorrect predictions after its first correct prediction (number of \textit{lapses}) and average the ratio of all intervals and over all questions.

DS metric is defined as:

\begin{equation}
\begin{aligned}
\textbf{DS}
&= \frac{1}{Q}\sum_{j=1}^{Q} \frac{1}{C}
   \sum_{k=1}^{N_j}\sum_{t=\tau_{k,j}+1}^{|T_{k, j}|}
   d_{j,k,t}, \\
d_{j,k,t}
&= \mathds{1}\!\left[p_{t,j} \neq a_{t,j}\right]
   \cdot \mathds{1}\!\left[\tau_{k,j} > 0\right].
\end{aligned}
\end{equation}

\paragraph{Phase Miss rate~(PM)}
PM measures the rate at which a model completely fails to capture the correct state throughout an entire phase. We quantify PM by summing up the duration of each phase where the model missed completely and averaging the ratio of all intervals over all questions.

PM is defined as:

\begin{equation}
    \textbf{PM} = \frac{1}{Q}\sum_{j=1}^{Q} \frac{1}{C} \sum_{k=1}^{N_j}|T_{k,j}|\cdot \mathds{1}[\tau_{k,j} = 0]
\end{equation}

\subsubsection{Statistical Reliability}
Due to the incremental nature of our evaluation, where models must perform inference at every interval, the total number of required inferences is exceptionally high. Specifically, for each model, the total inference count is calculated as $D \times C \times Q$, where $D$ represents the number of documents(books), $C$ the number of chunks per document, and $Q$ the number of questions per interval. This results in approximately 78k inferences for \Babi and 67k for \Novels per model. The baseline experiment on Qwen3-30B model required approximately 125 GPU hours on Nvidia A100 GPUs. Given the substantial computational cost, we report results from a single comprehensive run for each model rather than multiple trials with different seeds. However, the sheer volume of evaluation points across thousands of unique context-question pairs provides a high degree of consistency and statistical significance for our comparative analysis.

\section{Experimental Setup }
\label{app:exp_setup}

\subsection{Base Models}
\label{app:base-models}
We assess the proposed dataset on a total of 14 language models, including both open-source and proprietary systems, across a wide spectrum of model scales. Because long-context capacity and architectural design vary substantially across models, we summarize these characteristics in Table~\ref{tab:model_architecture}.

\begin{table*}[t!]
\centering
\fontsize{8}{10} \selectfont
\begin{tabular}{c  c l  c c c  c c  c}
\toprule
\multirow{2}{*}{\makecell{\textbf{Avail-} \\ \textbf{ability}}} & \multicolumn{2}{c}{\textbf{Model}} & \multicolumn{3}{c}{\textbf{Architecture}} & \multicolumn{2}{c}{\textbf{Context Length}} & \multirow{2}{*}{\makecell{\textbf{Exp.} \\ \textbf{Length}}} \\
\cmidrule(lr){2-3} \cmidrule(lr){4-6} \cmidrule(lr){7-8} 
& Model Family & Full-name & Size (Active) & Type & Attention & Default & Expanded & \\
\midrule

\multirow{12}{*}{Open} & \multirow{7}{*}{Qwen} & \href{https://huggingface.co/Qwen/Qwen3-4B-Instruct-2507}{Qwen3-4B-Instruct-2507} & 4B & Dense & GQA & 262k & - & 262k \\ 
 & & \href{https://huggingface.co/Qwen/Qwen2.5-7B-Instruct}{Qwen2.5-7B-Instruct} & 7B & Dense & GQA & 32k & 131k & 131k \\ 
 & & \href{https://huggingface.co/Qwen/Qwen3-8B}{Qwen3-8B} & 8B & Dense & GQA & 32k & 131k & 131k \\ 
 & & \href{https://huggingface.co/Qwen/Qwen3-30B-A3B-Instruct-2507}{Qwen3-30B-A3B-Instruct-2507} & 30B (3B) & MoE & GQA & 262k & - & 262k \\ 
 & & \href{https://huggingface.co/Qwen/Qwen3-30B-A3B-Thinking-2507}{Qwen3-30B-A3B-Thinking-2507} & 30B (3B) & MoE & GQA & 262k & - & 262k \\ 
 & & \href{https://huggingface.co/Qwen/Qwen3-Next-80B-A3B-Instruct}{Qwen3-Next-80B-A3B-Instruct} & 80B (3B) & MoE & Hybrid & 262k & 1M & 262k \\ 
 & & \href{https://huggingface.co/Qwen/Qwen3-235B-A22B-Instruct-2507}{Qwen3-235B-A22B-Instruct-2507} & 235B (22B) & MoE & GQA & 262k & 1M & 133k \\ 
\cmidrule{2-9}
 & \multirow{2}{*}{GPT-OSS} & \href{https://huggingface.co/openai/gpt-oss-20b}{gpt-oss-20b} & 20.9B (3.6B) & MoE & GQA & 131k & - & 131k \\ 
 & & \href{https://huggingface.co/openai/gpt-oss-120b}{gpt-oss-120b} & 116.8B (5.1B) & MoE & GQA & 131k & - & 131k \\ 
\cmidrule{2-9}
 & \multirow{2}{*}{Gemma} & \href{https://huggingface.co/google/gemma-3-4b-it}{Gemma 3-4b-it} & 4B & Dense & GQA & 131k & - & 131k \\ 
 & & \href{https://huggingface.co/google/gemma-3-27b-it}{Gemma 3-27b-it} & 27B & Dense & GQA & 131k & - & 131k \\ 
\midrule

\multirow{3}{*}{\makecell{Propr- \\ ietary}} & \multirow{2}{*}{Gemini} & Gemini 2.5 Flash & - & - & - & 1M & - & 1M \\ 
 & & Gemini 2.5 Pro & - & - & - & 1M & - & 1M \\ 
 & & Gemini 3.0 Pro & - & - & - & 1M & - & 1M \\ 
\bottomrule
\end{tabular}
\caption{Model Architecture and Performance Settings. Attention architecture of Qwen3-80B is a Hybrid of Gated DeltaNet and Gated Attention)} 
\label{tab:model_architecture}
\end{table*}

\subsubsection{RAG} 
Retrieval is restricted to chunks from previous time intervals, with chunk indices added to preserve order (i.e., no access to future information). 
Unless otherwise specified, we retrieve the total 30 relevant memory chunks from top, including both directly matched chunks and those linked through the framework's internal structure.

\subsubsection{Agentic Memory Systems}
HippoRAG-v2~\citep{gutierrez2025rag} is a graph-based retrieval framework built on Personalized PageRank. 
MemAgent~\citep{yu2025memagent} is designed for long-context processing with linear computational complexity, partitioning inputs into chunks and incrementally updating memory by combining prior memory with newly observed chunks at each timestamp, trained using GRPO.
A-Mem~\citep{xu2025mem} is an agentic memory system inspired by the Zettelkasten method and organizes memory as an interconnected knowledge network through dynamic indexing and bidirectional linking.

\subsubsection{Context Representation}
\paragraph{Baseline}
When evaluating on \novels, for models whose maximum sequence length is shorter than 1M tokens, we adopt a rolling-window strategy that always fills the model’s full context length, discarding the earliest chunks as new chunks are appended. For ultra-scale models such as Qwen3-235B, we cap the maximum model length at 133k tokens, since inference with the full length cannot be executed on a single 8 H100 node. The maximum context length used for each model in the baseline concatenation setting is reported in the final column of Table~\ref{tab:model_architecture}.

\paragraph{RAG}
To evaluate the models' tracking performance under a Retrieval Augmented Generation(RAG) setting, we structured the retrieval process and prompt formatting as follows:

\begin{enumerate}[leftmargin=1.5em, nosep, label=$\bullet$]
    \item \textbf{Indexing and Retrieval}: We build a vector index where each entry is formatted as "\texttt{\# chunk index : \{chunk\_idx\}, context: \{chunk\_text\}}". For a given question $q_j$ at interval $t$, we perform a similarity search using a query that incorporates the chunk index to maintain structural consistency with "\texttt{\# chunk index : \{chunk\_idx\}, question: \{question\}}"
    \item \textbf{Prompt Construction}: The top $k$ retrieved chunks are then prepended to the model's input prompt. To ensure the model can distinguish between different retrieved fragments, we use a structured list format as "\texttt{- Retrieved context:\textbackslash n\# chunk index : \{chunk\_idx\}, context: \{context\}\textbackslash n\# chunk index : \{chunk\_idx\}, context: \{context\}}"
\end{enumerate}

For RAG+R.W, the RAG component is only activated when the total available context exceeds the current window size ($t > w$), where $w$ denotes the window size, utilizing the historical chunks beyond the window for retrieval. The context is distinguished from each other with the format: "\texttt{- Retrieved context:\textbackslash n\# chunk index : \{chunk\_idx\}, context: \{context\}\textbackslash n\# chunk index : \{chunk\_idx\}, context: \{context\}\textbackslash n\textbackslash n \# Recent Chunks: \{context\}\textbackslash n\{context\} }" and the interval information is embedded into the question as well "\texttt{Current Head Index : \{chunk\_idx\}, question: \{question\}}"

\subsubsection{Inference Parameters}
\label{app:inference parameters}
For inference, we generally follow the best practices recommended by the model providers when available. As a default configuration, we adopt the settings used for the Qwen3 series: a temperature of 0.7, top-p of 0.8, and top-k of 20. Some models use provider-specified configurations that differ from this baseline, including Qwen3-Next-80B (temperature 0.6, top-p 0.95, top-k 20)
and Qwen3-30B-Thinking (temperature 0.6, top-p 0.95, top-k 20). For Gemini 2.5 Flash and Gemini 2.5 Pro, we use the default setup of temperature 0.0, top-p 0.95, and top-k 40. 
We generate up to 4096 tokens for all models. When querying Gemini with the thinking process enabled, we use the model's default mode, where it automatically budgets up to 8192 thinking tokens. For Gemini 3.0 Pro, we extended the maximum generation tokens to 32,768 in cases where the model exhausted the initial 4k-token limit in thinking mode without returning a final answer.

When the concatenated context exceeds a model’s default context length and the model supports context extension via YaRN, we expand the maximum sequence length accordingly, following the procedures specified in the official documentation.

Experiments were conducted using 4 or 8 A100 80G GPUs, 8 H100 80G GPUs, or 4 H200 140G GPUs based on the model size and availability.
\subsubsection{Inference Prompt}

The prompts used to evaluate models on \babi and \novels are provided in Prompt~\ref{infer_prompt_babi} and Prompt~\ref{infer_prompt_novel}, respectively. The task description is shared across the two benchmarks; however, the only difference lies in (i) how the “not answerable” option is defined and (ii) the output format. Specifically, \babi requires open-ended generation, whereas \novels requires selecting from predefined answer options. Our prompts are adapted from the original prompt used in the BABILong benchmark~\citep{kuratov2024babilong} and follow the best practice described in official document of Qwen3. 
To ensure deterministic evaluation, especially for models prone to divergent outputs, we appended a strict formatting constraint to the system prompt, where the model was instructed to conclude every response with a standardized template: "\texttt{\#\# Answer: [CHAR]}".

\begin{figure*}[t]
    \begin{promptcolorbox}[label=infer_prompt_babi]{Prompt}{Prompt for evaluation on \babi}{
    I will give you context and a question. You need to answer the question based only on the information from the text. \\
    
    Follow these rules carefully:
    \begin{enumerate}[leftmargin=1.5em, nosep, label=\arabic*.]
        \item Strict Contextual Grounding: You must base your reasoning exclusively on the text provided. Do not use any external or prior knowledge of this story (even if you recognize it).
        \item State Persistence: When a state is established (e.g., "The box is on the table"), you must assume that state persists unless a later part of the text explicitly changes or contradicts it (e.g., "...he moved the box into the basement.").
        \item Final State Priority: Your answer must always reflect the state of the subject at the end of the provided text. If a state changes multiple times, your answer must be based only on the most recent, final version.
        \item If the answer cannot be found in the context, respond with `Unknown'. If the question is a comparison and the values are same, respond with `Same'. Otherwise, give a concise response in 1–2 words (not a complete sentence).
        \item You may do reasoning, but in the end, output your final response in the format: `\#\# Answer: {short answer}'. The short answer will be parsed for an exact match.\\
        Example: If the question is `Where is Sandra?' and the answer is `kitchen', your response must end with `\#\# Answer: kitchen'. (Do not include extraneous text like `Sandra is in the kitchen').
    \end{enumerate}
    
    \vspace{1em}
    \#\# Context:
    \begin{Verbatim}
    {context}
    \end{Verbatim}
    
    \#\# Question: 
    \begin{Verbatim}
    {question}
    \end{Verbatim}
    
    }
    \end{promptcolorbox}
\end{figure*}

\begin{figure*}[t]
    \begin{promptcolorbox}[label=infer_prompt_novel]{Prompt}{Prompt for evaluation on \novels}{
    I will give you context and a question. You need to answer the question based only on the information from the text. \\
    
    Follow these rules carefully:
    \begin{enumerate}[leftmargin=1.5em, nosep, label=\arabic*.]
        \item Strict Contextual Grounding: You must base your reasoning exclusively on the text provided. Do not use any external or prior knowledge of this story (even if you recognize it).
        \item State Persistence: When a state is established (e.g., "The box is on the table"), you must assume that state persists unless a later part of the text explicitly changes or contradicts it (e.g., "...he moved the box into the basement.").
        \item Final State Priority: Your answer must always reflect the state of the subject at the end of the provided text. If a state changes multiple times, your answer must be based only on the most recent, final version.
        \item Answer Selection: You must choose your answer from the provided options. Do not generate your own answer. If the text does not contain the information to answer the question, you must select the option that says "We cannot answer this question at this point".
        \item Output Instructions: Please show your choice in the end of the answer field with only the choice letter, e.g., "answer": "C".
    \end{enumerate}
    
    \vspace{1em}
    \#\# Context:
    \begin{Verbatim}
    {context}
    \end{Verbatim}
    
    \#\# Question: 
    \begin{Verbatim}
    {question}
    \end{Verbatim}
    
    \#\# Options: 
    \begin{Verbatim}
    {options}
    \end{Verbatim}
    
    }
    \end{promptcolorbox}
\end{figure*}

\section{Evaluation Result}

\subsection{Relationship with reasoning/thinking ability}
Table~\ref{table: reasoning and thinking-novel} presents the performance results on \novels. Unlike the results observed in \babi, the activation of Thinking mode in \novels leads to a performance degradation. While \babi is a synthetic dataset specifically designed to require explicit multi-evidence integration, \novels necessitate implicit multi-hop reasoning woven into a naturalistic narrative. We hypothesize that inference-time scaling is most effective for tasks with high structural complexity, such as those requiring the simultaneous tracking of multiple states as seen in \babi, rather than the nuanced linguistic extraction required by \novels.

\subsection{Analysis on Performance of RAG and RW}
\label{app:rag-rw}

\begin{table}[t!]
\centering
\fontsize{8}{10} \selectfont
\setlength{\tabcolsep}{3pt} 
    \begin{tabular}{lcccccccc}
    \toprule
    & \multicolumn{4}{c}{\babi} & \multicolumn{4}{c}{\novels}  \\
    \cmidrule(lr){2-5} \cmidrule(lr){6-9}
    Strategy & \cellcolor{tablegray}All & Sprs. & Mod. & Freq.  & \cellcolor{tablegray}All & Sprs. & Mod. & Freq. \\
    \midrule
    Base &\cellcolor{tablegray}35.8 & 38.5 & 37.3 & 29.4 & \cellcolor{tablegray}62.8 & 71.2 & 63.7 & 57.1\\
    RAG & \cellcolor{tablegray}\textbf{37.8} & \textbf{40.1} & \textbf{39.2} & \textbf{32.0} & \cellcolor{tablegray}61.0 & 68.9 & 63.0 & 54.9\\
    R.W. & \cellcolor{tablegray}36.4 & 38.8 & 38.3 & 29.9 & \cellcolor{tablegray}62.6 & 69.2 & 63.6 & 58.0\\
    RAG+R.W & \cellcolor{tablegray}36.7 & 39.3 & 38.1 & 30.7 & \cellcolor{tablegray}\textbf{64.6} & \textbf{72.1} & \textbf{64.7} & \textbf{60.1}\\
    \bottomrule
    \end{tabular}
\caption
     {
     Accuracy (\%) of Qwen3-30B on \babi and \novels under different context construction strategies.
     We compare baseline concatenation, RAG with top-30 retrieved chunks, a rolling window of the most recent 30 chunks (RW), and their combination (RAG+RW; top-15 retrieved and 15 most recent).
     } 
\label{table: rag-rw}
\end{table}

\paragraph{No clear winner on optimal context construction strategies.}

Table~\ref{table: rag-rw} compares different context construction strategies under long-context constraints, including the Base setting, RAG, Rolling Window~(RW), which retains only the most recent 30 chunks, and their combination~(RAG+RW). 
While each strategy improves performance over the Base setting in most cases, no single approach consistently dominates across datasets. 
Specifically, RAG achieves the best performance on \babi (+2.0\%), suggesting that relevance-based evidence localization is effective when salient evidence is clearly distinguishable from background context. 
In contrast, on \novels, RAG degrades performance~(-1.8\%), whereas the combined RAG+RW strategy performs best~(+1.8\%).
We hypothesize that this is because novels are narrative-heavy contexts with complex temporal structure. Thus, even when retrieval is successful, models may struggle to effectively integrate retrieved passages, while the additional RW component preserves temporally aligned context, facilitating better understanding and utilization of accumulated long-term history~\citep{han2025rag,jimenez2024hipporag,lee2024well}.

\begin{figure}[t!]
    \centering
    \begin{minipage}[b]{\linewidth}
        \centering
            \centering
            \includegraphics[width=\linewidth]{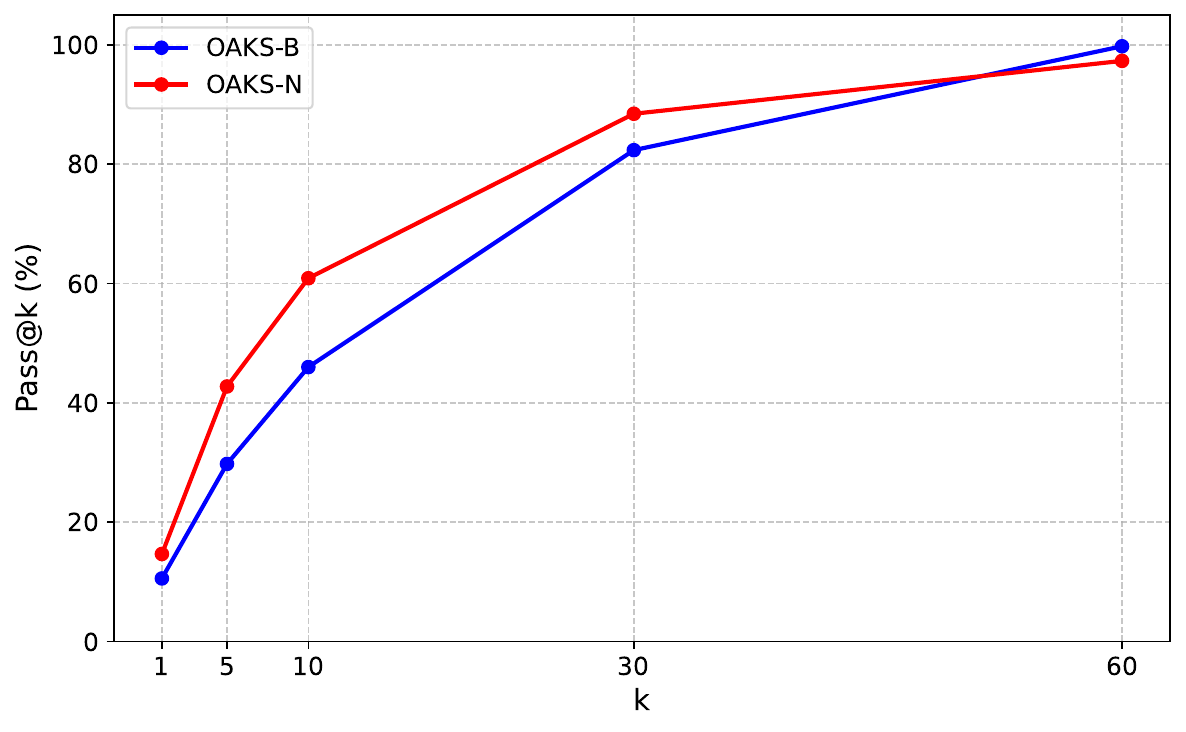}
        \caption{Pass@k performance as the number of context increases for retrieval in RAG on \babi and \novels}
        \label{fig:perf-by-RAG-k}
    \end{minipage}
\end{figure}

\paragraph{Performance by number of context chunks in RAG and Rolling Window}
\label{app:perf_rag_rw}
Figure~\ref{fig:perf-by-RAG-k} illustrates the performance of the retriever, Qwen3-Embedding-0.6B~\citep{qwen3embedding}, reporting the Pass@k metric as a function of the number of retrieved chunks. Given that our dataset necessitates the synthesis of multiple pieces of evidence, a single chunk rarely contains the complete ground truth. For the purpose of this analysis, however, we proxy the gold truth as the chunk of the state transition within the current phase. We observe that retriever performance increases monotonically, approaching near-perfect recall as $k$ reaches 60.

\begin{figure}[t!]
    \centering
    \begin{minipage}[b]{\linewidth}
        \centering
            \centering
            \includegraphics[width=\linewidth]{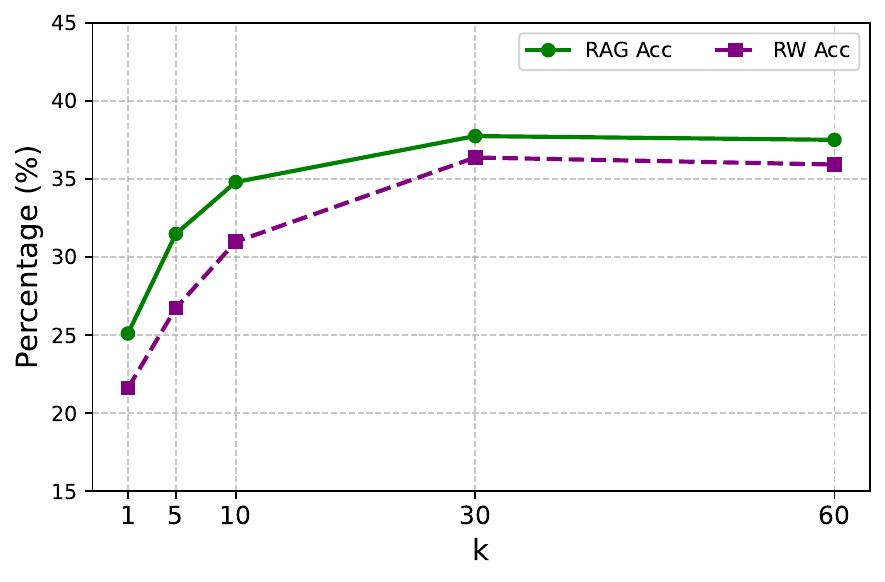}
        \caption{Accuracy~(\%) on \babi across different numbers of context chunks in RAG and RW. }
        \label{fig:perf-by-ragK-rwK}
    \end{minipage}
    \end{figure}

\begin{figure}[t!]
        \centering
    \begin{minipage}[b]{\linewidth}
        \centering
            \centering
            \includegraphics[width=\linewidth]{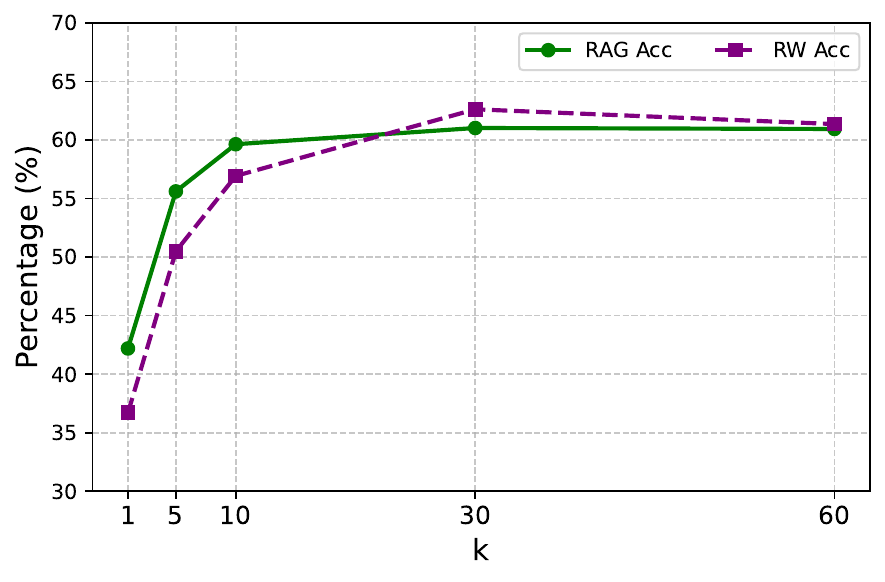}
        \caption{Accuracy~(\%) on \novels across different numbers of context chunks in RAG and RW. }
        \label{fig:perf-by-ragK-rwK-novel}
    \end{minipage}
\end{figure}

Figure~\ref{fig:perf-by-ragK-rwK} and Figure~\ref{fig:perf-by-ragK-rwK-novel} present the performance as a function of the number of chunks in RAG and RW on \babi and \novels, respectively. 
In both settings, increasing the number of chunks initially improves the performance but peaks around 30 chunks.

\begin{table}[t!]
\centering
\fontsize{8}{10}  
\selectfont
    \begin{tabular}{lc|cccc}
    \toprule
    Model & Think & All & Sparse & Moderate & Frequent  \\
    \midrule
    \multirow{2}{*}{\makecell[l]{Qwen3\\30B}} 
    & - & \textbf{62.8} & \textbf{71.2 }& \textbf{63.7} & \textbf{57.1}\\
    & \cmark  & 61.8 & 70.6 & 61.1 & 57.1\\

    \midrule
    \multirow{2}{*}{\makecell[l]{Gemini 2.5\\Flash} } 
    & - & \textbf{67.3} & \textbf{77.3} & \textbf{68.2} & \textbf{60.9}\\
    & \checkmark  & 65.6 & 75.2 & 65.6 & 60.0\\

    \bottomrule
    \end{tabular}
\caption{
Accuracy (\%) of Qwen3-30B and Gemini 2.5 with RAG on \novels by answer change frequency with and without Thinking Mode.
} 
\label{table: reasoning and thinking-novel}
\end{table}

\subsection{Difference by dataset}
\label{app:novel-vs-babi}
We observe that the average score for \babi~(39.4\%) is lower than that of \novels(57.5\%).
We conjecture that the multiple-choice format of \novels introduces a restricted answer space,\footnote{A random baseline can achieve an expected accuracy of 18.6\%.} and that the narratives within \novels may still be susceptible to the influence of prior knowledge bias. 
The lower performance on \babi may be attributed to its synthetic design, which allows for the creation of intrinsically more complex and difficult tasks.

\section{Analysis}
\label{app:Analysis}

\begin{table}[t!]
    \centering
    \fontsize{8}{10} \selectfont
        \begin{tabular}{lc|c}
        \toprule
        Models & Size& LongBench\\
        \midrule
        Qwen2.5 &7B& 30.0\\
        Gemma 3 &4B& 29.4\\
        GPT-OSS& 20B & 14.8\\
        Gemma 3 &27B & 33.6\\
        Qwen3 &30B & 32.6\\
        \bottomrule
        \end{tabular}
    \caption
         {Performance on LongBench-V2.} 
    \label{table: longbench}
\end{table}

\begin{figure*}[t]
    \begin{promptcolorbox}[label=llm_judge_evidence]{Prompt}{Prompt for LLM Judge over evidence}{
    
    You are an expert Evidence Quality Auditor. Your task is to verify if the Model's reasoning path matches the specific evidence logic found in the Ground Truth (GT) Evidence. \\
     \\
    \#\# Evaluation Data \\
    - **Question:** [INSERT QUESTION] \\ 
    - **GT Evidence:** [INSERT GT EVIDENCE] \\
    - **Model Output:** [INSERT MODEL OUTPUT] \\
     \\
    \#\# Evaluation Criteria \\
    - The "Core Anchor" Alignment: Identify the specific factual anchor in the GT Evidence (e.g., a specific location or name). The model must use this anchor as its primary basis for the answer. \\
    - Narrative Context Allowance: The model MAY use external knowledge of the source material (e.g., "In Chapter 1," "the inciting incident," "the parchment") to frame or organize its reasoning. This is seen as helpful context. \\
    - The "Source Primacy" Rule (Strict): The model MUST NOT use its own knowledge to contradict, correct, or bypass the GT Evidence. Phrases like "Based on my own knowledge, the text is wrong" or reaching a conclusion that ignores the GT facts in favor of external ones must be marked as FAIL. \\
    - Derivation Integrity: The model must explicitly link the location/fact back to the specific phrases found in the GT (e.g., "Konigstrasse in Hamburg"). It fails only if it reaches the answer using logic that ignores or replaces the GT Evidence entirely. \\
     \\
    \#\# Instructions \\
    - Extract GT Logic: Summarize the core reason the GT Evidence supports the answer. \\
    - Extract Model Logic: Summarize the core reason the Model provides in its reasoning. \\
    - Compare:  \\
    	- If the Model's logic is a direct reflection of the GT Evidence: PASS. \\
    	- If the Model's logic is different, uses outside facts, or misses the "Why" provided in the GT: FAIL. \\
     \\
    \#\# Response Format \\
    GT Logic Summary: <What is the specific 'Why' in the GT evidence?> Model Logic Summary: <What is the specific 'Why' in the model's reasoning?> Comparison: <Do they align? Does the model rely on the same sentences/logic?> \\
     \\
    Verdict: [PASS or FAIL]
    
    }
    \end{promptcolorbox}
\end{figure*}

\begin{figure*}[t]
    \centering
    \begin{minipage}[b]{\linewidth}
        \centering
            \centering
            \includegraphics[width=\linewidth]{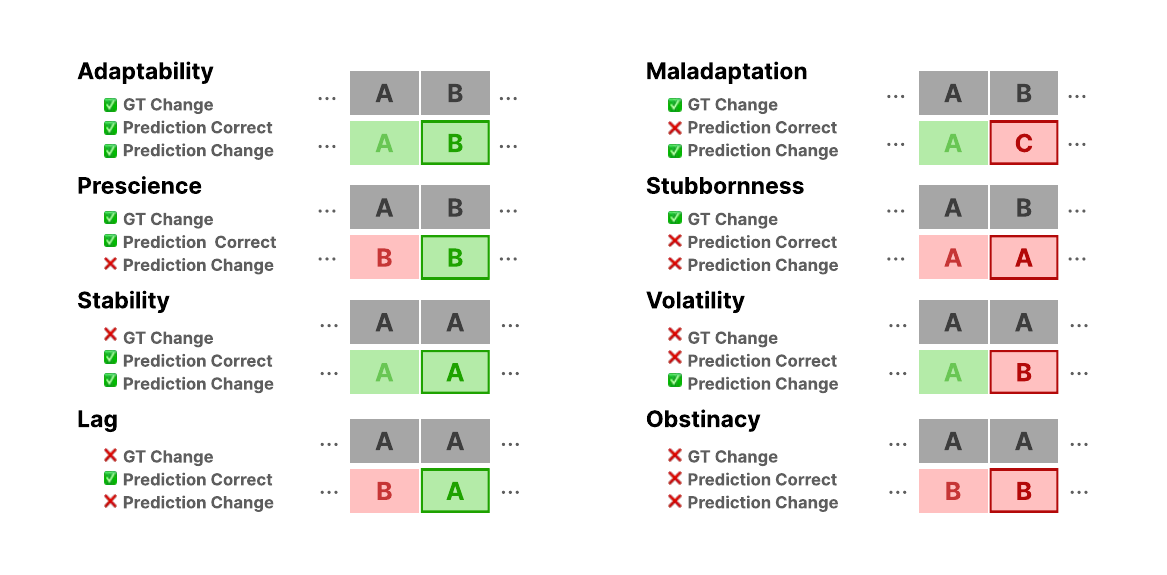}
        \caption{Extended visual explanations for model's behavior on \ours described in Table 6.}
        \label{fig:evidence_answer}
    \end{minipage}
\end{figure*}

\subsection{Fine-grained Analysis of Predicted Knowledge Transition Behavior}
\label{app:behavior}

In this section, we provide an additional detailed analysis based on Table~\ref{table: behavior} in Section~\ref{sec:behavior analysis}, which summarizes the results for representative models.
Table~\ref{table: full_behavior_babi} and Table~\ref{table: full_behavior_novel} show the analysis over \babi and \novels, respectively. Figure~\ref{fig:evidence_answer} shows a schematic illustration of model behaviors.

\paragraph{Correct transitions are easier than correct answers:}
Aggregating across models, we observe that models tend to correctly predict whether a transition occurs compared to producing the correct answer itself.
On average, correct transition behaviors occur more often (31.2\%) than incorrect ones (18.8\%), whereas answer correctness is lower overall (22.8\% correct vs. 27.2\% incorrect).
This indicates that identifying when to update knowledge is an easier subproblem than determining what the updated answer should be. 

\paragraph{Dominant behaviors when answer is correct?}
We analyze the model behavior among the cases where the answer is correct~(\textbf{\greencmark}). 
When the transition prediction is correct, \texttt{Adaptability} occurs more frequently than \texttt{Stability}, indicating that maintaining stable knowledge is more challenging than updating knowledge at the correct time. We hypothesize that this difficulty arises due to long and complex, multiple interacting facts within the given context, which increases the likelihood of spurious updates.

\paragraph{Dominant behaviors when answer is incorrect?}
We analyze the model behavior among the cases where the answer is incorrect~(\textbf{\redxmark}). 
In these cases, \texttt{Volatility} and \texttt{Maladaptation} are the more frequent behaviors when the transition is incorrect or correct, respectively, indicating that erroneous answers often coincide with unnecessary or poorly timed changes.
Compared to the correctly answered cases in the previous paragraph, the gap between changing and staying behaviors is smaller when the answer is incorrect (7.5 for incorrect vs. 14.6 for correct), suggesting that errors arise from a broader range of failure modes.

\subsection{Correct evidence prediction does not ensure correct answer prediction}
We conduct a detailed error analysis of the relation between answer and evidence prediction for \novels to evaluate how well the model uses the source from the book. We use an LLM as a judge to assess whether the predicted reasoning aligns with the annotated evidence and to check whether the model relies on its own knowledge.\footnote{Cases in which the model produced an answer directly without explicit reasoning were removed from this analysis}
The largest error category is when both answer and evidence are incorrect (47.3\%), indicating that failures frequently co-occur.
Fully correct predictions, where both answer and evidence are correct, account for 18.8\% of cases.
In 14.2\% of cases, the model produces the correct answer despite incorrect evidence, typically by relying on its own parametric knowledge or focusing on an incorrectly related passage. 
When evidence is correct, the model still produces an incorrect answer in 19.7\% of cases, often due to confusion among multiple answer options, highlighting that even correct evidence does not guarantee a correct answer and is consistent with the overall lower accuracy of answer prediction compared to evidence prediction. Taken together, these observations indicate that answer reasoning is a critical ability for solving this task.
We used Gemini 2.5 Pro as Judge. Prompt~\ref{infer_prompt_novel} shows the prompt used for LLM Judge.

\subsection{\ours is not solvable with simple long context understanding ability}

We analyze the correlation between performance on long context tasks and on \ours. 
To measure long-context ability, we use LongBench-v2~\citep{bai2024longbench2}, a widely adopted benchmark for evaluating models' performance on long context inputs~(Table~\ref{table: longbench}).
Across five models of similar size, the Pearson correlation between LongBench-v2 performance and performance on \babi and \novels is 0.69 and 0.34, respectively. 
When analysis over the \textit{Frequent} subset of \babi and \novels, the correlation drops to 0.45 and 0.30.
This suggests that while \ours benefits from a model's long context ability, as contexts of \ours are usually long, it also requires additional capabilities: adapting to dynamic knowledge online and accurately tracking evolving information.
Similarly, the analysis in Section~\ref{sec6:acc-context-length} shows that performance tends to degrade as the time interval increases, further highlighting that simple long-context understanding is insufficient to solve \ours.

\begin{table*}[t!]
\centering
\fontsize{8}{10} \selectfont
\setlength{\tabcolsep}{3pt} 
    \begin{tabular}{clccccccccc}
    \toprule
     &&&  \multicolumn{4}{c}{\text{GT Phase Transitions} (\underline{\textbf{C}}hange)} & \multicolumn{4}{c}{\text{No GT Transition} (\underline{\textbf{S}}tay)}\\
    \cmidrule(lr){4-7} \cmidrule(lr){8-11} 
    \multirow{2}{*}{Think} & \multirow{2}{*}{Models} & \multirow{2}{*}{Size} &  \multirow{2}{*}{\begin{tabular}[c]{@{}c@{}} \texttt{Adaptability} \\ (C / \greencmark) \end{tabular}}  
    & \multirow{2}{*}{\begin{tabular}[c]{@{}c@{}} \texttt{Maladaptation} \\ (C / \redxmark) \end{tabular}}
     & \multirow{2}{*}{\begin{tabular}[c]{@{}c@{}} \texttt{Prescience} \\ (S / \greencmark) \end{tabular}}  
     & \multirow{2}{*}{\begin{tabular}[c]{@{}c@{}} \texttt{Stubbornness} \\ (S / \redxmark) \end{tabular}}    
     & \multirow{2}{*}{\begin{tabular}[c]{@{}c@{}} \texttt{Lag} \\ (C / \greencmark) \end{tabular}}  
    & \multirow{2}{*}{\begin{tabular}[c]{@{}c@{}} \texttt{Volatility} \\ (C / \redxmark) \end{tabular}}
     & \multirow{2}{*}{\begin{tabular}[c]{@{}c@{}} \texttt{Stability} \\ (S / \greencmark) \end{tabular}}  
     & \multirow{2}{*}{\begin{tabular}[c]{@{}c@{}} \texttt{Obstinacy} \\ (S / \redxmark) \end{tabular}}  \\
    &  & && \\
    \midrule
    \multirow{12}{*}{\xmark}   
    & \multirow{5}{*}{Qwen3} 
    & 4B & 28.4 & 26.2 & 2.5 & 42.9 & 4.6 & 23.6 & 21.4 & 50.4 \\
    && 8B &  36.5 & 48.2 & 3.7 & 11.6 & 11.0 & 57.9 & 21.5 & 9.6 \\
    && 30B &  34.3 & 33.6 & 9.7 & 22.4 & 8.9 & 36.7 & 26.3 & 28.1 \\
    && 80B &   36.0 & 29.5 & 9.2 & 25.3 & 9.5 & 31.3 & 31.3 & 28.0 \\
    && 235B &  39.8 & 27.9 & 11.3 & 21.0 & 12.0 & 31.6 & 34.5 & 21.9 \\
    \cmidrule{2-11}
    & Qwen2.5 & 7B & 28.6 & 37.2 & 5.9 & 28.3 & 9.2 & 46.0 & 14.8 & 30.0 \\
    \cmidrule{2-11}
    & \multirow{2}{*}{GPT-OSS}
    & 20B & 33.3 & 54.6 & 3.8 & 8.2 & 9.3 & 68.5 & 12.0 & 10.2 \\
    && 120B & 39.1 & 40.8 & 5.5 & 14.6 & 12.8 & 46.5 & 24.2 & 16.5 \\
    \cmidrule{2-11}
    & \multirow{2}{*}{Gemma 3}
    & 4B & 26.5 & 44.0 & 7.4 & 22.1 & 7.9 & 48.7 & 15.5 & 27.9 \\
    && 27B & 31.6 & 28.6 & 11.9 & 27.9 & 12.2 & 27.0 & 25.2 & 35.6 \\
    \cmidrule{2-11}
    & \multirow{2}{*}{Gemini 2.5}
    & Flash &  36.3 & 17.0 & 17.2 & 29.5 & 7.5 & 16.9 & 35.0 & 40.7\\
    && Pro& 37.2 & 20.2 & 15.4 & 27.2 & 8.3 & 20.6 & 33.8 & 37.2 \\
    \midrule
    \multirow{3}{*}{\cmark} &  
    Qwen3&30B   & 39.6 & 34.6 & 7.7 & 18.2 & 13.0 & 37.7 & 30.4 & 18.9\\
    \cmidrule{2-11}
    & \multirow{2}{*}{Gemini 2.5}
     & Flash  & 47.5 & 23.8 & 15.0 & 13.7 & 12.4 & 27.5 & 43.3 & 16.8 \\
     && Pro&49.3 & 25.3 & 14.1 & 11.3 & 15.6 & 27.3 & 44.4 & 12.6 \\
    \bottomrule
    \end{tabular}
\caption{
Analysis of model knowledge tracking behavior on \babi. 
The table shows the average occurrence rate of specific tracking behaviors across all time intervals.
Results are partitioned by the ground truth (GT) state: whether the answer \textbf{Change} from the previous interval or \textbf{Stay} the same. 
Since Stay intervals are more frequent(94\% of all intervals), rates are averaged within each GT category to sum to 100\%.
The second row shows the behavioral labels (e.g., Maladaptation) and the model's action in the sub-header (e.g., (C / \greencmark)): whether the predicted answer \textbf{C}hanged from the previous prediction or \textbf{S}tay, and whether the resulting answer is correct (\textbf{\greencmark}) or wrong (\textbf{\redxmark}). 
}
\label{table: full_behavior_babi}
\end{table*}

\begin{table*}[t!]
\centering
\fontsize{8}{10} \selectfont
\setlength{\tabcolsep}{3pt} 
    \begin{tabular}{clccccccccc}
    \toprule
     &&&  \multicolumn{4}{c}{\text{GT Phase Transitions} (\underline{\textbf{C}}hange)} & \multicolumn{4}{c}{\text{No GT Transition} (\underline{\textbf{S}}tay)}\\
    \cmidrule(lr){4-7} \cmidrule(lr){8-11} 
    \multirow{2}{*}{Think} & \multirow{2}{*}{Models} & \multirow{2}{*}{Size} &  \multirow{2}{*}{\begin{tabular}[c]{@{}c@{}} \texttt{Adaptability} \\ (C / \greencmark) \end{tabular}}  
    & \multirow{2}{*}{\begin{tabular}[c]{@{}c@{}} \texttt{Maladaptation} \\ (C / \redxmark) \end{tabular}}
     & \multirow{2}{*}{\begin{tabular}[c]{@{}c@{}} \texttt{Prescience} \\ (S / \greencmark) \end{tabular}}  
     & \multirow{2}{*}{\begin{tabular}[c]{@{}c@{}} \texttt{Stubbornness} \\ (S / \redxmark) \end{tabular}}    
     & \multirow{2}{*}{\begin{tabular}[c]{@{}c@{}} \texttt{Lag} \\ (C / \greencmark) \end{tabular}}  
    & \multirow{2}{*}{\begin{tabular}[c]{@{}c@{}} \texttt{Volatility} \\ (C / \redxmark) \end{tabular}}
     & \multirow{2}{*}{\begin{tabular}[c]{@{}c@{}} \texttt{Stability} \\ (S / \greencmark) \end{tabular}}  
     & \multirow{2}{*}{\begin{tabular}[c]{@{}c@{}} \texttt{Obstinacy} \\ (S / \redxmark) \end{tabular}}  \\
    &  & && \\
    \midrule
    \multirow{12}{*}{\xmark}   
    & \multirow{5}{*}{Qwen3} 
    & 4B & 51.8 & 15.2 & 7.5 & 25.5 & 8.2 & 20.4 & 38.6 & 32.8 \\
    && 8B &  53.3 & 18.4 & 6.7 & 21.7 & 8.2 & 23.1 & 43.1 & 25.7 \\
    && 30B &  54.8 & 9.7 & 11.9 & 23.7 & 5.4 & 10.5 & 57.9 & 26.2 \\
    && 80B &   55.5 & 8.5 & 12.5 & 23.5 & 4.6 & 8.7 & 60.3 & 26.4 \\
    && 235B &  55.2 & 9.0 & 11.3 & 24.5 & 5.0 & 9.5 & 60.3 & 25.2 \\
    \cmidrule{2-11}
    & Qwen2.5 & 7B & 27.9 & 39.1 & 4.2 & 28.9 & 12.2 & 31.1 & 19.7 & 37.0 \\
    \cmidrule{2-11}
    & \multirow{2}{*}{GPT-OSS}
    & 20B & 48.2 & 23.0 & 5.8 & 23.0 & 10.7 & 33.5 & 33.6 & 22.2 \\
    && 120B & 46.8 & 23.2 & 8.3 & 21.7 & 11.0 & 23.4 & 43.4 & 22.3 \\
    \cmidrule{2-11}
    & \multirow{2}{*}{Gemma 3}
    & 4B & 37.2 & 14.5 & 16.0 & 32.2 & 4.5 & 14.0 & 32.7 & 48.9 \\
    && 27B & 57.2 & 8.0 & 12.3 & 22.5 & 5.0 & 9.4 & 56.0 & 29.6 \\
    \cmidrule{2-11}
    & \multirow{2}{*}{Gemini 2.5}
    & Flash &  59.9 & 10.4 & 9.8 & 19.9 & 6.3 & 12.8 & 61.4 & 19.6 \\
    && Pro& 25.4 & 35.2 & 1.2 & 38.2 & 15.7 & 27.0 & 9.4 & 47.8 \\
    \midrule
    \multirow{3}{*}{\cmark} &  
    \multirow{2}{*}{Gemini 2.5}
     & Flash  & 60.8 & 15.7 & 9.7 & 13.8 & 8.5 & 19.5 & 56.7 & 15.4 \\
     && Pro&67.3 & 6.2 & 13.9 & 12.7 & 4.1 & 7.5 & 72.6 & 15.8 \\
    \bottomrule
    \end{tabular}
\caption{
Analysis of model knowledge tracking behavior on \novels. 
}
\label{table: full_behavior_novel}
\end{table*}

\section{Ethical Considerations / Potential Risks}
Our study is primarily restricted to English narratives, which may limit the generalizability of our findings to other languages.
\ours incurs substantial computational cost, which would indicate increased energy consumption.
We hired human annotators when constructing \novels; although we applied extensive filtering and verification procedures, potential annotation errors may remain due to the inherent subjectivity of human judgment.

\section{Usage of LLM}
In our research, we employed Large Language Models~(LLMs) for initial data generation and writing assistance. In \novels, as explained in Section~\ref{sec3: dataset} and Appendix~\ref{app:novels}, initial drafts of question-answer pairs were generated by Gemini 2.5 Pro, which were subject to rigorous human annotation process to filter out low-quality questions. During writing, LLMs were utilized for sentence-level refinement and grammatical polishing. All AI-generated suggestions were carefully reviewed and edited by the authors to maintain the coherency and accuracy.

\end{document}